\documentclass[11pt]{article}
\usepackage[margin=1.1in]{geometry}

\newcommand{\customfont}{\fontfamily{Gyre Pagella}\selectfont}
\DeclareTextFontCommand{\textcomicneue}{\customfont}

\newcommand{\comicneue}{\fontfamily{ComicNeue-TLF}\selectfont}
\DeclareTextFontCommand{\textcomicneue}{\comicneue}

\usepackage[T1]{fontenc}

\usepackage{natbib}

\usepackage{graphicx}
\usepackage{booktabs}
\usepackage{array}

\usepackage{hyperref}

\usepackage{longtable}

\usepackage{color}
\usepackage{booktabs}       
\usepackage{amsfonts}       
\usepackage{nicefrac}       
\usepackage{microtype}      
\usepackage{lipsum}
\usepackage{bbm}
\usepackage{dsfont}

\usepackage{pgffor}
\usepackage{bbm}
\usepackage{graphicx}
\usepackage{subcaption}
\usepackage{multirow}
\usepackage[normalem]{ulem}

\usepackage{amsmath}
\newcommand{\okedit}[1]{#1}

\newcommand{\mgedit}[1]{#1}
\newcommand{\mgminor}[2]{#2}

\newif\ifappendix

\appendixtrue

\newcommand{\comm}[1]{\textcolor{cyan}{}}

\newcommand{\B}[1]{\mathbf{#1}}
\usepackage{amsthm}

\newtheorem{definition}{Definition}

\theoremstyle{definition}
\newtheorem{example}{Example}[section]

\title{Neural Markov Logic Networks}
\author{%
   Giuseppe Marra \\
   Department of Computer Science \\
   KU Leuven \\
   Leuven, Belgium \\
   \and
   Ond\v{r}ej Ku\v{z}elka \\
   Faculty of Electrical Engineering \\
   Czech Technical University in Prague \\
   Prague, Czech Republic \\
}

\begin{document}

\maketitle

\begin{abstract}
    We introduce neural Markov logic networks (NMLNs), a statistical relational learning system that borrows ideas from Markov logic. Like Markov logic networks (MLNs), NMLNs are an exponential-family model for modelling distributions over possible worlds, but unlike MLNs, they do not rely on explicitly specified first-order logic rules. Instead, NMLNs learn an implicit representation of such rules as a neural network that acts as a potential function on fragments of the relational structure. 
    Similarly to many neural symbolic methods, NMLNs can exploit embeddings of constants but, unlike them, NMLNs work well also in their absence. This is extremely important for predicting in settings other than the transductive one. We showcase the potential of NMLNs on knowledge-base completion, triple classification and on generation of molecular (graph) data.
\end{abstract}

\section{INTRODUCTION}


Statistical relational models are typically learned from one or more examples of relational structures that typically consist of a large number of ground atoms. Examples of such structures are social networks,
protein-protein interaction networks etc. A challenging task is to learn a probability distribution over such relational structures from one or few examples. One possible approach is based on the assumption that the relational structure has repeated regularities; this assumption is implicitly or explicitly used in most works on statistical relational learning.
Statistics about these regularities can be computed for small substructures of the training examples and used to construct a distribution over the \mgminor{}{whole} relational structures. Together with the maximum-entropy principle, this leads to exponential-family distributions such as Markov logic networks \citep{richardson2006markov}. In classical MLNs, however, either domain experts are required to design some useful statistics about the domain of interest by hand (i.e.\ logical rules) or they need to be learned by structure learning based on combinatorial search. \mgminor{There was also recently a flurry of new methods trying to improve relational learning with neural computation}{Recently, many authors have tried to improve relational learning by integrating it with neural computation} \citep{rocktaschel2017end,DBLP:conf/aaai/Kazemi018,DBLP:journals/jair/SourekAZSK18}. However, these hybrid approaches usually relax (or drop) the goal of modeling the joint probability distribution, preventing them from being applied to more complex learning and reasoning tasks.

In this paper, we propose neural Markov logic networks (NMLN). Here, the statistics (or features), which are used to model the probability distribution, are not known in advance, but are modelled as neural networks trained together with the probability distribution model. NMLNs overcome several limitations of existing approaches. In particular, \textit{(i)} they can be used as an out-of-the-box tool in heterogeneous domains; \textit{(ii)} they allow expressing and learning joint probability distributions of complete relational structures. 

The main contributions \okedit{presented in this paper} are as follows:
\textit{(i)} \mgminor{}{we introduce a new class of potential functions exploiting symmetries of relational structures}; 
\textit{(ii)} \mgminor{}{we introduce a new statistical relational model called neural Markov logic networks and we propose a theoretical justification of the model as emerging from a principle of Min-Max-entropy}; 
\textit{(iii)}  \mgminor{}{we identify subclasses of NMLNs that allow for faster inference}; 
\textit{(iv)}  \mgminor{}{we showcase the model's effectiveness on three diverse problems: knowledge-base completion, triple classification and generative modelling of small molecules.}

\mgminor{}{The paper is structured as follows. In Section \ref{sec:preliminaries}, we introduce preliminary concepts. In Section \ref{sec:potentials} we define relational potential functions. In Section \ref{sec:nmlns}, we introduce the NMLN model. In Section \ref{sec:exps}, we show the results of the experiments we conducted. In Section \ref{sec:related}, we position the proposed model in the literature. Finally, in Section \ref{sec:conclusions}, we draw some conclusions.}



\section{PRELIMINARIES}
\label{sec:preliminaries}


We consider a function-free first-order logic language $\mathcal{L}$, which is built from a set of constants $\mathcal{C}_\mathcal{L}$ and predicates $\mathcal{R}_\mathcal{L} = \bigcup_i \mathcal{R}_i$, where $\mathcal{R}_i$ contains the predicates of arity i. 
For $c_1, c_2, \dots, c_m \in \mathcal{C}_\mathcal{L}$ and $R\in\mathcal{R}_m$, we call $R(c_1, c_2, \dots, c_m)$ a \textit{ground atom}.
We define \textit{possible world} $\omega$ to be the pair $(\mathcal{C}, \mathcal{A})$, where $\mathcal{C} \subseteq \mathcal{C}_\mathcal{L}$, $\mathcal{A}$ is a subset of the set of all ground atoms that can be built from the constants in $\mathcal{C}$ and any relation in $\mathcal{R}_\mathcal{L}$. We define 
$\Omega_\mathcal{L}$ to be the set of all possible worlds over $\mathcal{L}$. 
Intuitively, a given possible world defines a set of $true$ facts one can state using the constants (entities) and the relations of the language~$\mathcal{L}$.

\begin{definition}[Fragments]
Let $\omega = (\mathcal{C},\mathcal{A})$ be a possible world. A \textit{fragment} $\omega \langle \mathcal{S} \rangle$ is defined as the restriction
of $\omega$ to the constants in $\mathcal{S}$. It is a pair $\omega \langle S \rangle = (\mathcal{S}, \mathcal{B})$, with $\mathcal{S}$ the constants of the restriction and $\mathcal{B}$ a set of ground atoms only using constants from~$\mathcal{S}$.  
\end{definition}


\begin{example}\label{example1} Given a language based on the set of constants $\mathcal{C}_\mathcal{L} = \{Alice, Bob, Eve\}$ and a set of relations $\mathcal{R}_\mathcal{L} = \{sm(x), fr(x,y)\}$, consider a possible world on this language:
\[
\omega = (\mathcal{C}_\mathcal{L}, \{sm(Alice), fr(Alice,Bob), fr(Bob,Eve)\}).\]

Then, for instance, the fragment induced by the set of constants 
$\mathcal{S} = \{Alice, Bob\}$ is 
\[\omega \langle S \rangle = (\mathcal{S}, \{sm(Alice), fr(Alice,Bob)\}).\]
\end{example}

The set of all fragments of $\omega$ that are induced by size-$k$ subsets of constants will be denoted by $\Gamma_{k}(\omega)$. Similarly, $\Gamma_k(\mathcal{L})$ will denote the set of all possible fragments in a given first-order language $\mathcal{L}$.

\section{RELATIONAL POTENTIALS}
\label{sec:potentials}

A fundamental ingredient of neural Markov logic networks is the careful design of the potential function of the distribution. It needs to represent and exploit symmetries in the possible worlds to correctly shape the distribution. In this section, \mgminor{we provide the definitions and the design choices needed to introduce the relational potentials which will be then instantiated using neural networks.}{we define and design relational potentials. We also show how to instantiate them using neural networks.} 

\subsection{Fragment and Global Potentials}

We need two classes of relational potentials: fragment potentials and global potentials, which are defined on fragments and possible worlds, respectively.

\begin{definition}[Fragment Potential]
Given a first-order logic language $\mathcal{L}$, a fragment potential function $\phi$ is any parametric function $\phi(\gamma; \mathbf{w})$ from $\Gamma_k(\mathcal{L})$ to $\mathbb{R}$ with a parameter vector $\mathbf{w}$.
\end{definition}

\begin{definition}[Global potential]
Given a parametric fragment potential function $\phi(\gamma;\mathbf{w})$, we define the parametric global potential function:
$$\Phi(\omega; \mathbf{w}) = \frac{1}{|\Gamma_k(\omega)|} \sum_{\gamma \in \Gamma_k(\omega)} \phi(\gamma; \mathbf{w}).$$
\end{definition}

\subsection{Symmetric Fragment Potentials}

In order to find and exploit relational symmetries in possible worlds, we need to introduce the concept of symmetric potentials. 
Symmetric potentials are useful \mgminor{when we want}{} to model relational structures, such as molecules, where the same molecule \mgminor{may be represented in different ways but at the same time isomorphic molecules are physically indistinguishable}{may have many isomorphic variants, which are physically indistinguishible}. In general, symmetric potentials are also useful for learning models that can be applied on examples containing different sets of constants (e.g.\ when learning from multiple social networks rather than one).
 
\begin{definition}[Symmetric fragment potentials]
A fragment potential function $\phi$ is symmetric if $\phi(\gamma, \mathbf{w}) = \phi(\gamma',\mathbf{w})$ whenever $\gamma$ and $\gamma'$ are isomorphic.
\end{definition}

\mgminor{}{The term} {\em isomorphic} above means that $\gamma$ can be obtained from $\gamma'$ by renaming (some of) the constants (here, the renaming must be an  injective mapping). \mgminor{Clearly, i}{I}f the fragment potential $\phi$ is symmetric then the global potential $\Phi$ must be symmetric as well, i.e.\ if $\omega$ and $\omega'$ are isomorphic possible worlds then $\Phi(\omega;\mathbf{w}) = \Phi(\omega';\mathbf{w})$.

\mgminor{To represent symmetric potential functions, we need the concept of {\em fragment anonymization}.}{Once we have defined symmetric fragment potentials, we still need to represent them. To this end, we use the concept of {\em fragment anonymization}.} Given a fragment $\gamma = (\mathcal{S},\mathcal{B})$, its anonymizations $\operatorname{Anon}(\gamma)$ is a list of $|\mathcal{S}|!$ fragments obtained as follows. First, we construct the set of all bijective mappings from the set $\mathcal{S}$ to $\{1,2,\dots, |\mathcal{S}| \}$. \mgminor{We call this set}{This is the set of} {\em anonymization functions} and \mgminor{}{we} denote it by $\operatorname{AnonF}(\gamma)$.
\mgminor{Each of the $|\mathcal{S}|!$ anonymizations is then obtained }{Then, we obtain each of the elements of $\operatorname{Anon}(\gamma)$} by taking one function from $\operatorname{AnonF}(\gamma)$ and applying it to $\gamma$. Note that $\operatorname{Anon}(\gamma)$ may contain several identical elements.

\begin{example}\label{ex:anon1}
Consider again the fragment $\gamma = (\mathcal{S}, \{sm(Alice), fr(Alice,Bob)\})$. The set of anonymization functions is $\operatorname{AnonF}(\gamma) = \{ \{\textit{Alice} \mapsto 1, \textit{Bob} \mapsto 2 \}, \{ \textit{Alice} \mapsto 2, \textit{Bob} \mapsto 1 \} \}$ and the respective list of anonymizations is then the list: 
$\operatorname{Anon}(\gamma) = ( \gamma', \gamma'')$ where $\gamma' = (\{1,2\},$ $\{sm(1),$$ fr(1,2)\})$ and $\gamma'' = (\{1,2\}, \{sm(2), fr(2,1)\})$.
\end{example}

We use anonymizations to define symmetric fragment potentials \mgminor{}{starting} from not necessarily symmetric functions. Specifically, for a given function $\phi'(\gamma;\mathbf{w})$ on fragments over the language $\mathcal{L}_0$ with $\mathcal{C}_{\mathcal{L}_0} = \{1,2,\dots,k\}$, we define the symmetric fragment potential as 
\begin{equation}
    \phi(\gamma; \mathbf{w}) =   \sum_{\gamma' \in \operatorname{Anon}(\gamma)} \phi'(\gamma'; \mathbf{w}),
    \label{eq:symmetric_potential}
\end{equation}
Clearly, any potential computed as above must be symmetric (and, vice versa,\mgminor{it is not difficult to show that}{} any symmetric potential can be represented in this way).

\subsection{Neural Fragment Potentials}
Anonymizations of fragments can also be represented using binary vectors. All possible ground atoms that we can construct from the available relations (from $\mathcal{R}_\mathcal{L}$) and the constants from $\{1,2,\dots,|\mathcal{S}| \}$ can be ordered (e.g. lexicographically) and then used to define the binary-vector representation.

\begin{example}\label{ex:anon2}
Let $\mathcal{R}_\mathcal{L} = \{\textit{sm}(x), \textit{fr}(x,y) \}$. Consider the fragment 
$\gamma$ from Example \ref{ex:anon1}. If we order the possible ground atoms lexicographically as: $\textit{fr}(1,1)$, $\textit{fr}(1,2)$, $\textit{fr}(2,1)$, $\textit{fr}(2,2)$, $\textit{sm}(1)$, $\textit{sm}(2)$, its two anonymizations can be represented by the binary vectors $(0,1,0,0,1,0)$ and $(0,0,1,0,0,1)$.
\end{example}

From now on we will treat anonymizations and their binary-vector representations interchangeably as long as there is no risk of confusion. Representing anonymizations as binary vectors allows us to represent the functions $\phi'$ above using standard \textit{feedforward neural networks}; the parameters $\mathbf{w}$ are then the weights of the neural network. These networks take, as input, a binary-vector representation of the current anonymization and return a real value as output. \mgminor{It is worth noticing that, w}{W}hen functions $\phi'$ are represented as neural networks, Equation \ref{eq:symmetric_potential} is actually defining a sharing scheme of the weights for the fragment potential $\phi$. This scheme is imposing, by construction, an invariance property w.r.t. isomorphisms of fragments. One can get a nice intuition about the properties of this class of functions when comparing them with Convolutional Neural Networks (CNN). \mgminor{While a CNNs an input and its spatial translation are mapped to the same set of features}{While a CNN computes the same set of features for an input and its spatial translation} (i.e. translation invariance), a symmetric fragment potential computes the same set of features for symmetric fragments.

\subsection{General Fragment Potentials}

Now we explain how to represent general non-symmetric potentials that will \mgminor{among others}{} allow us to learn vector-space embeddings of the constants, which is also a key feature of many existing transductive models, like NTP \citep{rocktaschel2017end}. In what follows \mgminor{the embedding parameters will be denoted by $\mathbf{W}_e$}{$\mathbf{W}_e$ will denote the embedding parameters} and $\mathbf{W}_e(c_1,\dots,c_k)$ will \mgminor{be}{denote} the concatentation of the embedding vectors of constants $c_1$, $\dots$,~$c_k$.

Consider a potential $\phi'(\gamma;\mathbf{w}, \mathbf{W}_e)$ on fragments over the language $\mathcal{L}_0$ with $\mathcal{C}_{\mathcal{L}_0} = \{1,2,\dots,k\}$ where $\mathbf{w}$ and $\mathbf{w}_e$ are some parameter vectors. As was the case for the symmetric potentials, using the binary-vector representation of fragments in $\mathcal{L}_0$, $\phi'$ can be represented, for instance, as a feedforward neural network. We can then write the general potential function as
\begin{align}
    \phi(\gamma; \mathbf{w}, \mathbf{W}_e) =  \sum_{\pi \in \operatorname{AnonF}(\gamma)} \phi'(\pi(\gamma); \mathbf{w}, \mathbf{W}_e(\pi^{-1}(1), \dots, \pi^{-1}(k)))
\end{align}

Again it is not difficult to show that any symmetric or non-symmetric fragment potential function can be represented in this way. We may notice that when $\mathbf{W}_e(c)$ gives the same vector for all constants, the potential will also be symmetric, which is not the case in general. As we show in Section~\ref{subsec:kbc}, the addition of embedding of constants helps improving the prediction capability of our model in transductive settings.

\section{NEURAL MARKOV LOGIC NETWORKS}
\label{sec:nmlns}


In this section we introduce {\em neural Markov logic networks} (NMLNs), an exponential-family model for relational data that is based on the potentials described in the previous section. 

\paragraph{Neural Markov Logic Networks}
Given a set of fragment potential functions $\phi_1$, $\dots$, $\phi_m$, and the respective global potential functions $\Phi_1$, $\dots$, $\Phi_m$, a \textit{neural Markov logic network (NMLN)} is the parametric exponential-family distribution over possible worlds from given $\Omega_{\mathcal{L}}$:
\[P(\omega) = \frac{1}{Z} \exp{\left(\sum_{i} \beta_i \Phi_i(\omega; \mathbf{w}_i, \mathbf{W}_e) \right)},\]
where $\beta_i$, $\mathbf{w}_i$ and $\mathbf{W}_e$ are parameters (when the potentials are symmetric, we drop $\mathbf{W}_e$) and $Z = \sum_{\omega \in \Omega_\mathcal{L}} \exp{\left(\sum_{i} \beta_i \Phi_i(\omega; \mathbf{w}_i, \mathbf{W}_e) \right)}$ is the normalization constant (partition function).

\subsection{First Observations}

Here we note several basic properties of NMLNs. First, an NMLN is symmetric \mgminor{, meaning that it gives the same probabilities to isomorphic worlds,}{if its  potential functions are symmetric}. \mgminor{}{A symmetric NMLN gives the same probabilities to isomorphic worlds}. Second, potentials in NMLNs play similar role as rules in classical MLNs. In fact, as we show in Appendix B, any MLN without existential quantifiers can be straightforwardly represented as an NMLN.\footnote{By using max-pooling in the definition of global potentials, one can obtain even richer class of NMLN models that can represent any ``Quantified MLN'' \citep{DBLP:conf/kr/Gutierrez-Basulto18a}. These models are not considered in this paper.} Third, standard first-order logic rules can be added simply as other potentials to NMLNs, which allows using expert-crafted rules next to learned neural potentials.



\subsection{The Min-Max Entropy Problem}

As we show in this section, NMLNs naturally emerge from a principle of min-max entropy. For simplicity, we assume that we have one training example $\widehat{\omega}$ (the same arguments extend easily to the case of multiple training examples). The example $\widehat{\omega}$ may be, for instance, a knowledge graph (represented in first-order logic). We want to learn a distribution based on this example.\footnote{Regarding consistency of learning SRL models from a single example, we refer e.g.\ to \citep{OndrejUai19}.} For that we exploit the principle of min-max entropy.

\paragraph{Maximizing Entropy} Let us first assume that the potentials $\Phi_i(\omega; \B w_i, \mathbf{W}_e)$ and the parameter vectors $\mathbf{w}_1$, $\dots$, $\mathbf{w}_m$, and $\mathbf{W}_e$ are fixed. Using standard duality arguments for convex problems \citep{boyd2004convex,wainwright2008graphical}, one can \mgminor{then}{} show that 
the solution of the following convex optimization problem is an NMLN: 
\begin{align}
& \underset{P_\omega}{\text{max}}
 - \sum_\omega P_\omega \log P_\omega  \\
& \text{s.t.}\quad
\sum_\omega P_\omega = 1, \; \forall \omega \colon p_\omega \ge 0 \;  \label{constr1}\\
& \quad\quad\forall i \colon \mathbb{E}_{P_\omega}[\Phi_i(\omega; \mathbf{w}_i, \mathbf{W}_e)] = \Phi_i(\widehat \omega; \mathbf{w}_i, \mathbf{W}_e)   \label{constr2}\; 
\end{align}

\mgminor{}{Here, the probability distribution is represented through the variables $P_\omega$, one for each possible world.}
This problem asks to find a maximum-entropy distribution such that the expected values of the potentials $\Phi_i$ are equal to the values of the same potentials on the training examples $\widehat{\omega}$, i.e.\ $\mathbb{E}_{P_\omega}[\Phi_i(\omega; \mathbf{w}_i, \mathbf{W}_e)] = \Phi_i(\widehat \omega; \mathbf{w}_i, \mathbf{W}_e)$. One can use Lagrangian duality to obtain the solution in the form of an NMLN: 
\[P_\omega = \frac{1}{Z} \exp{\left(\sum_{i} \beta_i \Phi(\omega; \mathbf{w}_i,\mathbf{W}_e)) \right)}.\]

Here, the parameters $\beta_i$ are solutions of the dual problem

\[\text{max}_{\beta_i}\; \{ \sum_{i=1}^M \beta_i \Phi_i(\widehat \omega; \mathbf{w}_i, \mathbf{W}_e)) - \log Z\},\]

which coincides with maximum-likelihood when the domain size of the training example $\widehat{\omega}$ and the domain size of the modelled distribution are equal.\footnote{We note that the derivation of the dual problem follows easily from the derivations in \citep{kuvzelka2018relational}, which in turn rely on standard convex programming derivations from \citep{boyd2004convex,wainwright2008graphical}. Throughout this section we assume that a positive solution exists, which is needed for the strong duality to hold; this is later guaranteed by adding noise during learning.} 
For details we refer to the discussion of model A in \citep{kuvzelka2018relational}.

\paragraph{Minimizing Entropy} Now we lift the assumption that the weights $\mathbf{w}$ and $\mathbf{W}_e$ are fixed. We learn them by minimizing the entropy of the max-entropy distribution. Let us denote by $H(\beta_1,\dots,\beta_m, \mathbf{w}_1, \dots, \mathbf{w}_m, \mathbf{W}_e)$ the entropy of the distribution $P_\omega$. We can write the resulting \textit{min-max} entropy problem as:
\begin{align}
    \nonumber \underset{\mathbf{w}_i,\mathbf{W}_e}{\text{min}}\;\underset{\beta_i}{\text{max}} H(\beta_1,\dots,\beta_M, \mathbf{w}_1, \dots, \mathbf{w}_m,\mathbf{W}_e) = \\ \nonumber -\underset{\mathbf{w}_i,\mathbf{W}_e}{\text{max}}\;\underset{\beta_i}{\text{min}} -H(\beta_1,\dots,\beta_M, \mathbf{w}_1, \dots, \mathbf{w}_m,\mathbf{W}_e)
\end{align}
\noindent subject to the constraints (\ref{constr1}) and~(\ref{constr2}).
Plugging in the dual problem and using strong duality, we obtain the following unconstrained optimization problem which is equivalent to the maximization of log-likelihood when the domain size of the training example $\widehat{\omega}$ and that of the modelled distribution are the same, i.e. the optimization problem is:
\begin{align}
\label{eq:max_likelihood}
\max_{\mathbf{w}_i,\mathbf{W}_e, \beta_i} \left\{ \sum_{i=1}^m \beta_i \Phi_i(\widehat \omega; \mathbf{w}_i, \mathbf{W}_e) - \log Z \right\}. 
\end{align}

The maximization of the log-likelihood will be carried out by a gradient-based method (see Appendix~A
).

\paragraph{Justification} 
Selecting the potential in such a way as to decrease the (maximum) entropy of MaxEnt models can be shown to decrease the KL-divergence between the true data distribution and the model distribution \citep{zhu1997minimax}. Of course, this coincides with maximum-likelihood estimation, but only when the domain size of the training example $\widehat{\omega}$ and that of the modelled distribution are the same; we refer to \citep{kuvzelka2018relational} for a more thorough discussion of this distinction for max-entropy models. Importantly, the min-max entropy perspective allows us to see the different roles that are played by the parameters $\beta_i$ on the one hand and $\mathbf{w}_i$ and $\mathbf{W}_e$ on the other.

\subsection{Inference}
\label{subsec:inference}


We use Gibbs Sampling (GS) for inference in NMLNs. 
Gibbs Sampling requires a large number of steps before converging to the target distribution. 
However, when we use it for learning inside SGD to approximate gradients  (Appendix~A
), we run it only for a limited number of steps (\citet{hinton2002training}).

\paragraph{Handling Determinism} 
Gibbs sampling cannot effectively handle distributions with determinism. In normal Markov logic networks, sampling from such distributions may be tackled by an algorithm called MC-SAT \citep{poon2006sound}. However, MC-SAT requires an explicit logical encoding of the deterministic constraints, which is not available in NMLNs where deterministic constraints are implicitly encoded by the potential functions.\footnote{In fact, only constraints that are almost deterministic, i.e.\ having very large weights, can occur in NMLNs but, at least for Gibbs sampling, the effect is the same. Such distributions would naturally be learned in our framework on most datasets.} Our solution 
is to simply avoid learning distributions with determinism by adding noise during training. We set a parameter $\pi_n \in [0,1]$ and, at the beginning of each training epoch, each ground atom of the input (training) possible worlds is inverted with probability $\pi_n$. This added noise also prevents the model from perfectly fitting training data, acting as a regularizer \citep{bishop1995training}.


\subsection{Faster Inference for $k \leq 3$}

The performance of Gibbs sampling can be improved using the idea of blocking \citep{jensen1995blocking} in combination with the massive parallelism available through GPUs. We describe such a method for NMLNs with fragments of size $k \leq 3$. For simplicity we assume that all the relations are unary or binary (although it is not difficult to generalize the method to higher arities). The description below applies to one pass of Gibbs sampling over all the possible ground atoms.

\paragraph{Case $k = 1$:} This is the most trivial case. Any two atoms $U(c)$, $U'(c')$ and $R(c,c')$ are independent when $c \neq c'$, where $U$, $U'$ and $R$ are some relations (in fact, all $R(c,c')$ where $c \neq c'$ have probability $0.5$ in this case). Hence, we can run in parallel one GS for each domain element $c$ -- each of these parallel GS runs over the unary atoms of the form $U(c)$ and reflexive binary atoms $R(c,c)$ for the given constant $c$ over all relations.

\paragraph{Case $k = 2$:} The sets of unary and reflexive atoms that were independent for $k=1$ are no longer independent. Therefore we first run GS sequentially for all unary and reflexive atoms. Conditioned on these, the atoms in any collection $R_1(c_1,c_1')$, $\dots$, $R_n(c_n,c_n')$ are independent provided $\{c_i,c_i'\} \neq \{c_j,c_j'\}$ for all $i \neq j$. Therefore we can now create one GS chain for every $2$-fragment and run these GS in parallel.\footnote{We can further increase scalability of NMLNs for $k=2$ for transductive-learning problems such as knowledge graph completion by exploiting {\em negative-based sampling}. Here, when using GS to estimate the gradients while training, instead of running it on all pairs of constants, we use only those pairs for which there are at least some relations and only a sub-sample of the rest of pairs (and estimate gradients from these, taking into account the subsampling rate).} We note that very similar ideas have been exploited in lifted sampling algorithms, e.g.\ \citep{venugopal2012lifting}.

\paragraph{Case $k = 3$:} We first sample unary and reflexive atoms as we did for $k=2$. Conditioned on these, the atoms in any collection $R_1(c_1,c_1')$, $\dots$, $R_n(c_n,c_n')$ are independent provided $\{c_i,c_i'\} \cap \{c_j,c_j'\} = \emptyset$ for all $i \neq j$ (compare this to $k=2$). This gives us a recipe for selecting atoms that can be sampled in parallel. First, we construct a complete graph of size $n$ and identify the constants from the domain with its vertices. We then find an edge-coloring of this graph with the minimum number of colors. When $n$ is even then $n-1$ colors are sufficient, when it is odd then we need $n$ colors (finding the coloring is trivial). We then partition the set of pairs of constants based on the colors of their respective edges. GS can be run in parallel for $2$-fragment corresponding to edges that have the same color. This may in principle bring an $O(n)$ speed-up (as long as the parallel GS chains fit in the GPU).\footnote{Another speed-up for transductive problems such as knowledge-graph completion, where we only care about ranking individual atoms by their marginal probabilities, can be achieved as follows. We sample subsets of the domain of size $m < n$. For each of the samples we learn an NMLN and predict marginal probabilities of the respective atoms using Gibbs sampling. At the end, we average the marginal probabilites.}

\section{EXPERIMENTS}
\label{sec:exps}

In this section, we report experiments done with NMLNs on three diverse tasks: knowledge base completion and triple classification on the one hand and molecular (graph) generation on the other. The aim of these experiments is to show that NMLNs can be used as an out-of-the-box tool for statistical relational learning. 

We implemented neural Markov logic networks in
Tensorflow\footnote{Downloadable link: provided with supplemental material during the review process.}. 
In Appendix
D, we provide detailed information about neural network architectures, hyperparameters grids and selected hyperparameters .

\subsection{Knowledge Base Completion}
\label{subsec:kbc}
In Knowledge Base Completion (KBC), we are provided with an incomplete KB and asked to complete the missing part. 
The KBC task is inherently in the transductive setting 
and the data are provided in a positive-only fashion: 
we cannot distinguish between unknown and false facts. \citet{OndrejUai19} studied KBC tasks under the missing-completely-at-random assumption and showed consistency of learning MLNs by maximum-likelihood where both missing and false facts are treated in the same way as {\it false}. Their arguments can be modified to give similar consistency guarantees for NMLNs.

\paragraph{Smokers.}
The ``Smokers'' dataset \citep{richardson2006markov} is a classical example in statistical relational learning literature. Here, two relations are defined on a set of constants representing people: the unary predicate \textit{smokes} identifies those people who smoke, while the binary predicate \textit{friendOf} indicates that two people are friends. This dataset is often used to show how a statistical relational learning algorithm can model a distribution by finding a correlation of smoking habits of friends. For example, in MLNs, one typically uses weighted logical rules such as: $\forall x \ \forall y \ \texttt{friendOf}(x,y) \rightarrow \texttt{smokes}(x) \leftrightarrow \texttt{smokes}(y)$.

\begin{figure}[t]
\centering
\begin{subfigure}{0.5\linewidth}
  \centering
  \includegraphics[width=\linewidth]{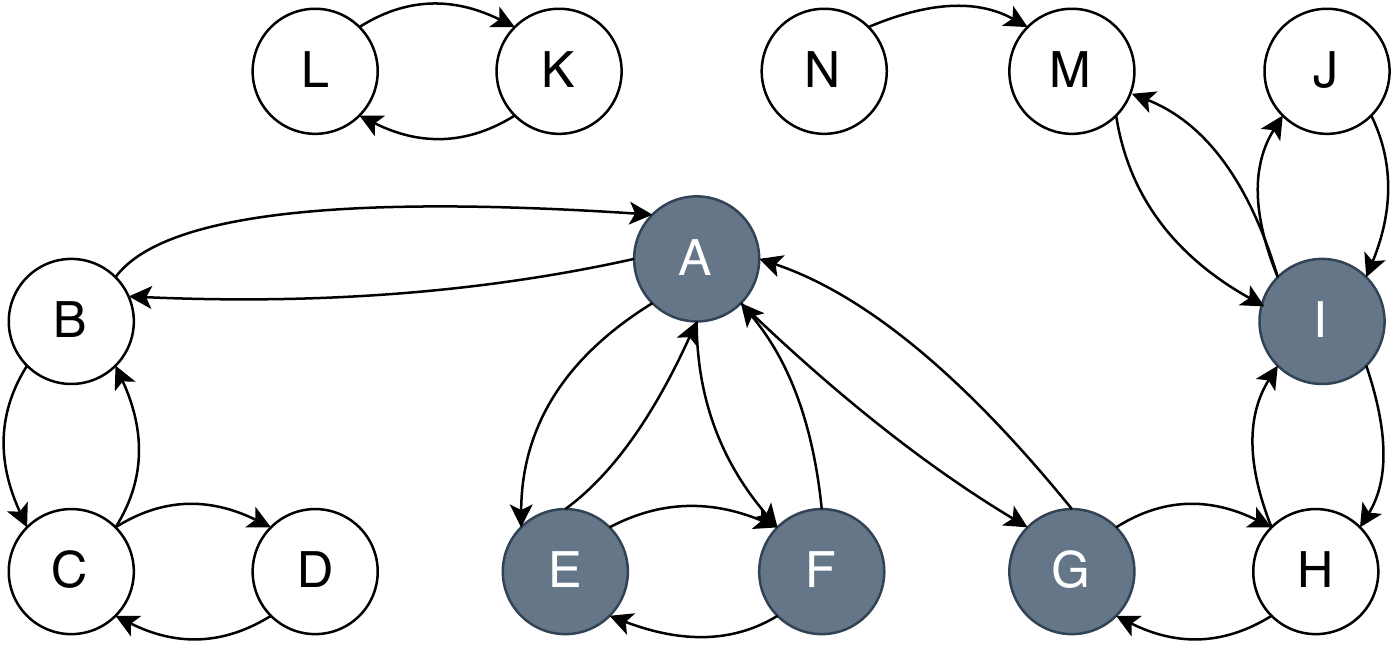}
  \caption{The training KB.}
  \label{fig:smokers_a}
\end{subfigure}
\hfill
\begin{subfigure}{0.5\linewidth}
  \centering
  \includegraphics[width=\linewidth]{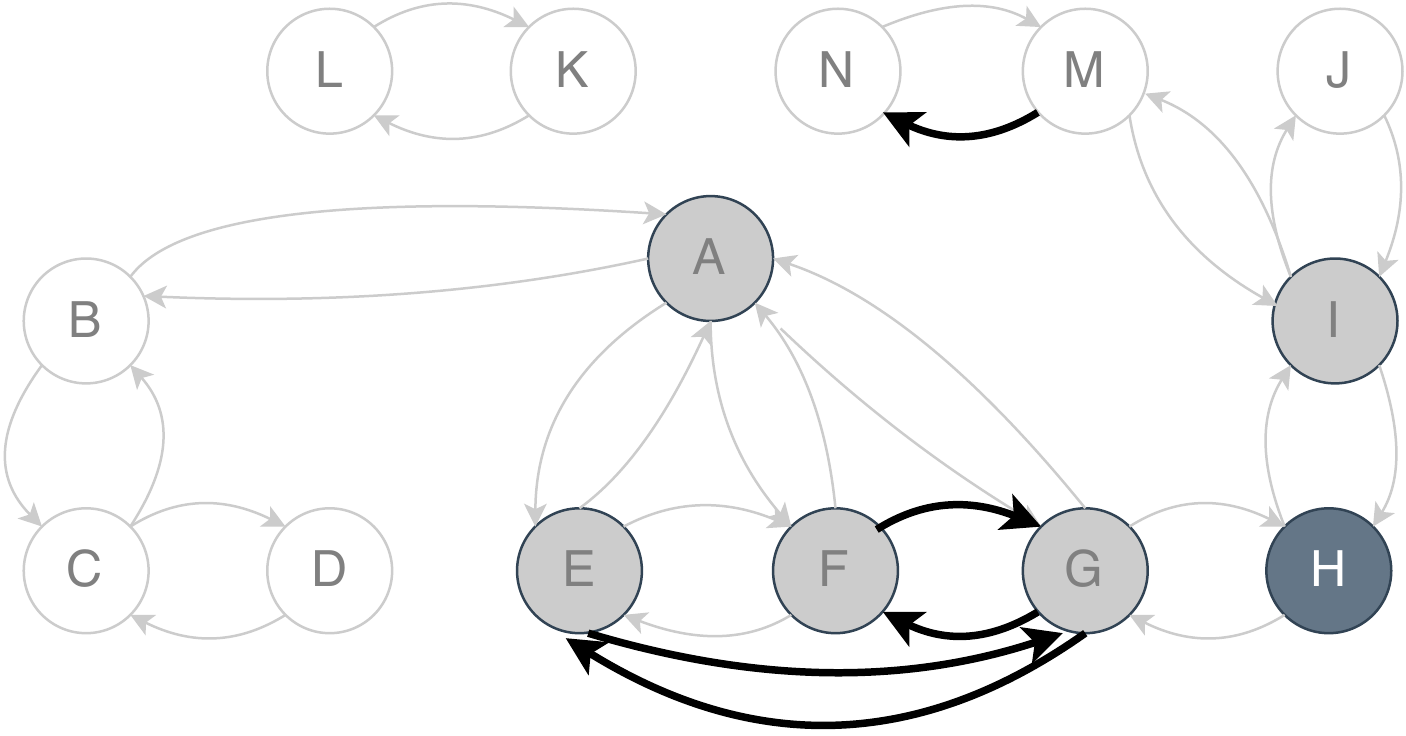}
  \caption{The completed KB.}
  \label{fig:smokers_b}
\end{subfigure}
\caption{\textbf{Knowledge Base Completion in the Smokers dataset.} 
A grey circle means that the predicate \texttt{smokes} is \textit{True} and a white one that it is unknown. Links represent the relation \texttt{friendOf}. 
The given world is shown in \ref{fig:smokers_a} and the completed one in \ref{fig:smokers_b}. The NMLN learnt that \texttt{friendOf} is symmetric, 
that a friend of at least two smokers is also a smoker, 
and that two smokers, who are friends of the same person, are also friends.
}
\label{fig:smokers}
\end{figure}

We trained a NMLN on the small smokers dataset. Since no prior knowledge about the type of rules that are relevant was used by NMLNs, the model itself had to identify which statistics are mostly informative of the provided data by learning the potential functions. We started here with the Smokers dataset in order to (i) illustrate the Knowledge Base Completion task and (ii) to provide some basic intuitions about what kind of rules the model could have learned.
In Figure~\ref{fig:smokers}, we show the KB before and after completion. In Figure~\ref{fig:smokers_b}, we highlight only new facts whose marginal probability after training is significantly higher than the others.

\paragraph{Nations, Kinship and UMLS} We use the Nations, Kinship and Unified Medical Language System (UMLS) KBs from \citep{kok2007statistical}. Nations contains 56 binary
predicates, 14 constants and 2565 true facts, Kinship contains 26 predicates,
104 constants and 10686 true facts, and UMLS contains 49 predicates, 135 constants and 6529 true
facts. These datasets have been exploited to test KBC performances in Neural Theorem Provers \citep{rocktaschel2017end}. 
\mgedit{Greedy-NTPs \citep{minervini2019differentiable} and CTP \citep{minervini2020learning} were recently introduced and the authors showed that their models were able to outperform original NTPs as well as other models proposed in the literature for tackling KBC tasks. In this section, we show how we can use NMLNs to tackle a KBC task on the Nations, Kinship and UMLS datasets.}


We followed the evaluation procedure in \citep{minervini2019differentiable}\footnote{The authors provided a corrected evaluation protocol and we exploited their evaluations routines}. In particular, we took a test fact and corrupted its first and second argument in all possible ways such that the corrupted fact is not in the original KB. Subsequently, we predicted a ranking of every test fact and its corruptions to calculate MRR and HITS@m. The ranking is based on marginal probabilities estimated by running Gibbs sampling on the NMLN. 
We compare the original Neural Therem Prover (\textit{NTP}) model \citep{rocktaschel2017end}, Greedy NTP (\textit{G-NTP}) \citep{minervini2019differentiable}, CTP \citep{minervini2020learning}, \textit{NMLNs} with $k = 3$ and no embeddings (\textit{NMLN-K3}) 
and NMLNs with $k=2$ and embeddings of domain elements (\textit{NMLN-K2E}). 
In Table~\ref{tab:kbc_nations}, we report the results of the KBC task. \textit{NMLN-K2E} outperforms competitors by a large gap on almost all datasets and metrics. Moreover, we can make two observations. Embeddings seem to play a fundamental role in the sparser datasets (i.e. Kinship and UMLS), where the relational knowledge is limited. However, both on Nations and Kinship, \textit{NMLN-K3} still performs better than differentiable provers, even if it cannot exploit embeddings to perform reasoning and it has to rely only on the relational structure of fragments to make predictions. 
This is a clear signal that, in many cases, the relational structure already contains a lot of information and that NMLNs are better in modeling and exploiting these relational regularities.

\begin{table*}[t]
  \centering
  \caption{ MRR and HITS@$m$ on Nations, Kinships and UMLS.}
  \label{tab:results}      
  \small{
  \begin{tabular}{llcccccc}
    \toprule
     \multirow{ 2}{*}{Dataset} & \multirow{ 2}{*}{Metric} & \multicolumn{5}{c}{Model}\\
    \cmidrule(lr){3-7}
     & & \multicolumn{1}{c}{\textbf{NTP}} & \multicolumn{1}{c}{\textbf{GNTP}} & 
     \multicolumn{1}{c}{\textbf{CTP}} & 
    \multicolumn{1}{c}{\textbf{NMLN-K3}} & 
    \multicolumn{1}{c}{\textbf{NMLN-K2E}}\\
    \midrule
    \multirow{4}{*}{Nations} 
    & MRR & $0.61$ & $0.658$ &  $0.709$& $\textbf{0.813}$ & $0.782$ \\
    & HITS@1 & $0.45$ & $0.493$ & $0.562$ & $\textbf{0.713}$
    & $0.667$ \\
    & HITS@3 & $0.73$ & $0.781$ & $0.813 $ & $\textbf{0.895}$ 
    &
    $0.863$  \\
    & HITS@10 & $0.87$ & $0.985$ & $0.995$ & $0.982$  
    
    & $\textbf{0.995}$ \\
    \midrule
    \multirow{4}{*}{Kinship} 
    & MRR & $0.35$ & $0.759$ & $0.764$ & $0.820$ & $\textbf{0.844}$  \\
    & HITS@1 & $0.24$ & $0.642$& $0.646$&  $0.737$ & $\textbf{0.769}$  \\
    & HITS@3 & $0.37$ & $0.850$ & $0.859$ & $0.886$ & $\textbf{0.903}$  \\
    & HITS@10 & $0.57$ & $0.959$ & $0.958$& $0.963$ & $\textbf{0.977}$  \\
    \midrule
     \multirow{4}{*}{UMLS} 
     & MRR & $0.80$ & $0.857$ & $0.852$& $0.502$ & $\textbf{0.924}$  \\
    & HITS@1 & $0.70$ & $0.761$ &$0.752$ & $0.391$  & $\textbf{0.896}$  \\
    & HITS@3 & $0.88$ & $\textbf{0.947}$ & $0.947$ & $0.549$ & $\textbf{0.940}$  \\
    & HITS@10 & $0.95$ & $\textbf{0.986}$ & $0.984$ & $0.713$ & $0.978$  \\
    \bottomrule
  \end{tabular}
  }
  \label{tab:kbc_nations}
\end{table*}

\subsection{Triple Classification}\label{sec:exp:kge} 
In triple classification, one is asked to predict if a given triple belongs or not to the knowledge base.  Even though there exists an entire class of methods specifically developed for this task, we wanted to show that our method is general enough to be also applicable to this large scale problems. We performed experiments on WordNet \citep{miller1995wordnet} and FreeBase \citep{bollacker2008freebase}, which are standard benchmarks for large knowledge graph reasoning tasks. We used the splits WN11 and FB13 provided in \citet{socher2013reasoning}. Interestingly, NMLN with $k=2$ achieves an accuracy of \textit{$(74.4,84.7)$} in WN11 and FB13, respectively. This compares similarly or favourably w.r.t. standard knowledge graph embeddings methods, like SE \citep{bordes2011learning} \textit{$(53.0, 75.2)$} or TransE \citep{bordes2012joint}$(75.9, 81.5)$. However, it is outperformed by newer methods, like  TransD \citep{ji2015knowledge} $(86.4, 89.1)$ and DistMult-HRS \citep{zhang2018knowledge} $(88.9, 89.0)$. This is likely due to the fact that these two datasets are extremely sparse and very few pairs of constants are related by more that one relation. Unlike knowledge graph embedding methods, which are tuned for the specific task of predicting head or tail entity of a triple, our model is general and learns a joint probability distribution. The fact that it can still perform similarly to state-of-the-art methods on this specialized task is in fact rather surprising.

\subsection{Graph Generation}
\label{subsec:mols_generation}

\begin{figure}
\centering
\includegraphics[width=0.9\linewidth, trim=0 0 0 0]{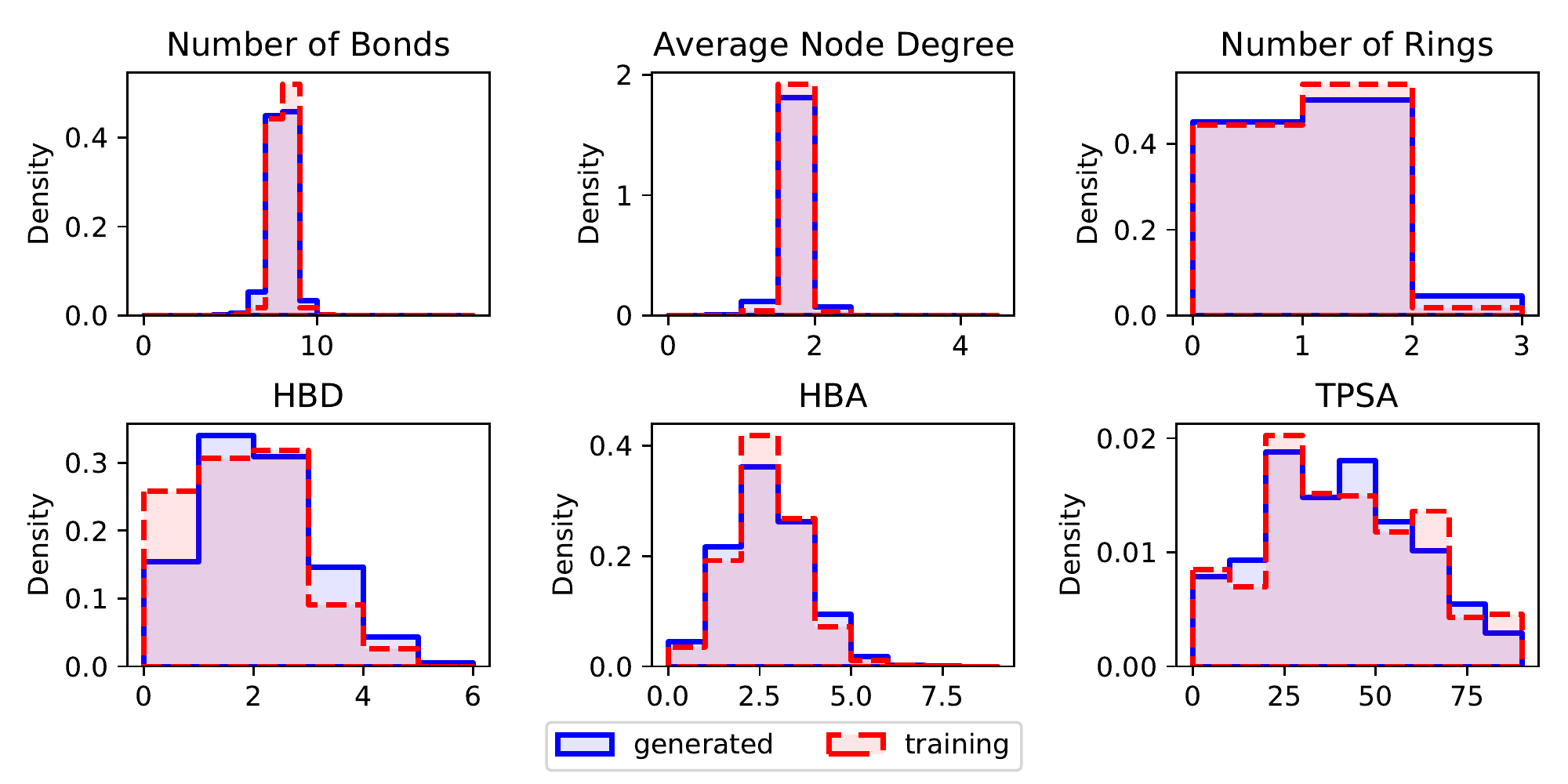}
\caption{\textbf{Molecules generation.} Comparison of chemical properties of generated and real molecules.}
\label{fig:stats}
\end{figure}

By modeling the joint probability distribution of a relational structure and by learning the potentials as neural networks, NMLNs are a valid candidate for generative tasks in non-euclidean settings, which are receiving an increasing interest recently \citep{you2018graphrnn,li2018learning}. To generate a set of relational structures, we can just collect samples generated by Gibbs sampling during training of an NMLN and return top-$n$ most frequently occurring ones (or, alternatively, top $n$ from the last $N$ sampled ones to allow time for NMLNs to converge). In this section, we describe a molecule generation task. We used as training data the ChEMBL molecule database \citep{gaulton2016chembl}. We restricted the dataset to molecules with 8 heavy atoms (with a total of 1073 training molecules). We used the RDKit framework \footnote{\texttt{https://rdkit.org/}} to get a FOL representation of the molecules from their SMILES encoding. 
We show a more detailed description of the training data and generation setting in Appendix 
C. In Figure~\ref{fig:molecules_main}, we show the set of top-$20$ sampled molecules. The first three molecules turn out to be isomers of {\em hexanoic acid}, the fourth is known as {\em 4-hydroxyvaleric acid}, the fifth is the alcohol called {\em heptanol} etc. 
Furthermore, in Figure~\ref{fig:stats}, we compare the normalized count of some statistics on the training and generated molecules, as it has been recently done in \citet{li2018learning}. These statistics represent both general structural properties as well as chemical functional properties of molecules, e.g.\ the topological polar surface area (TPSA) is an indicator of the capability of a molecule to permeate a membrane. 

\begin{figure}
    \centering
    \includegraphics[width=0.6\linewidth]{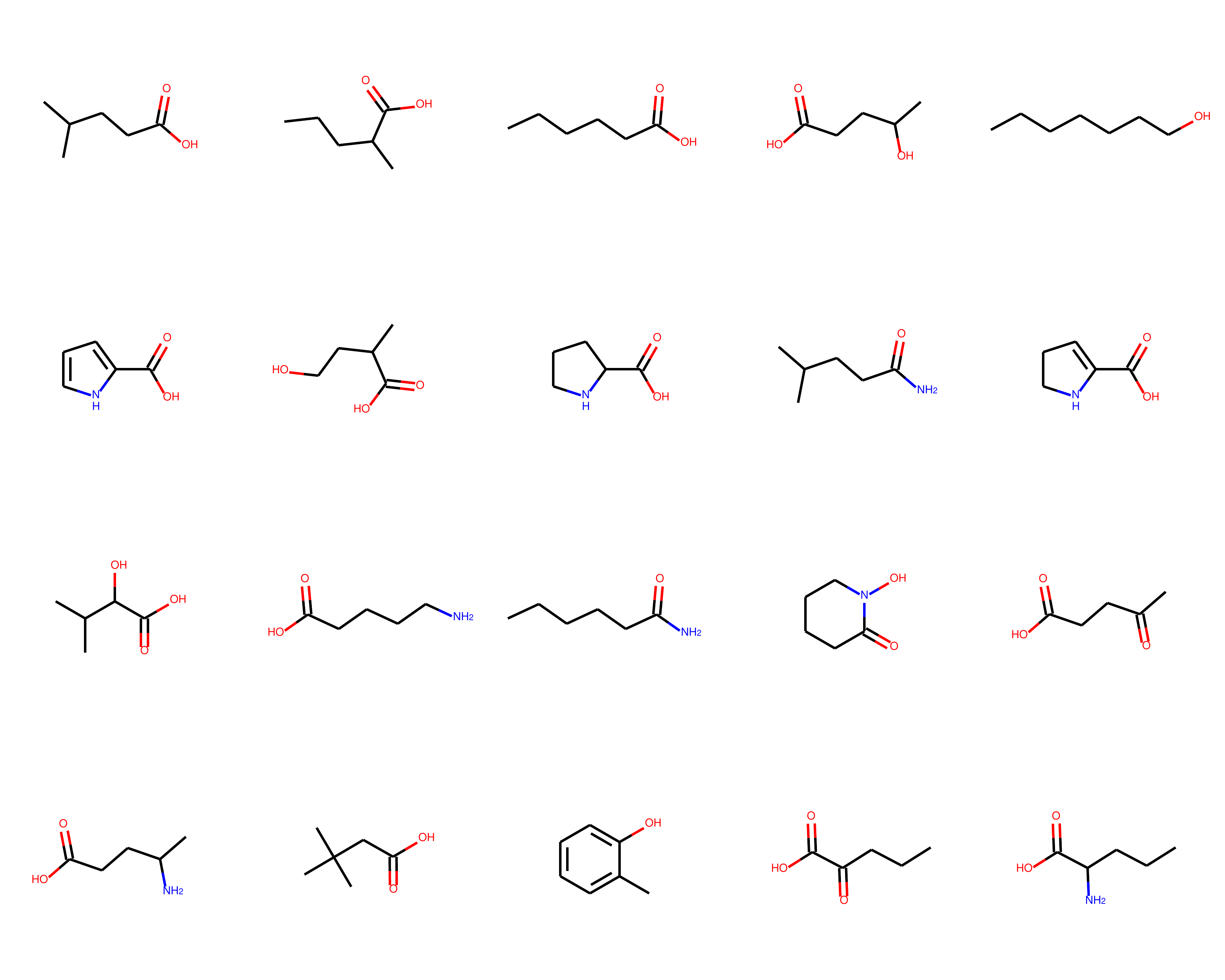}
    \caption{\textbf{Molecules generation.} A sample of generated molecules.}
    \label{fig:molecules_main}
\end{figure}


\section{RELATED WORK}
\label{sec:related}

\paragraph{NMLNs as SRL} NMLNs are an SRL framework inspired by Markov Logic Networks \citep{richardson2006markov}. 
From certain perspective, NMLNs can be seen as MLNs in which first-order logic rules are replaced by neural networks (for an explicit mapping from MLNs to NMLNs, refer to Appendix
B). While this may seem as a rather small modification, the gradient based learning of NMLNs' potentials allows more efficient learning than the usual combinatorial structure-learning.\footnote{We also briefly tried to learn a generative model for molecules in MLNs, but with absolutely no success. One of the reasons for this is likely that the number of logical rules needed for such a model is prohibitively large, whereas (an approximation) of such rules can be represented rather compactly in the neural potentials of NMLNs.} 
An alternative approach to improve performance of structure learning in MLNs is represented by the gradient boosted MLNs \citep{khot2015gradient}. However, as also noted in \citep{khot2015gradient}, the boosting approach has not been extended to the generative learning setting where one optimizes likelihood rather than pseudo-likelihood. In contrast, NMLNs are generative and trained by optimizing likelihood. Finally, unlike NMLNs, standard MLNs do not support embeddings of domain elements.

\paragraph{NMLNs as NeSy}
NMLNs integrate logical representations with neural computation, which is the domain of interest of Neural Symbolic Artificial Intelligence -- NeSy \citep{besold2017neural}. 
\citet{lippi2009prediction} was an early attempt to integrate MLNs with neural components. Here, an MLN was exploited to describe a conditional distribution over ground atoms, given some features of the constants. In particular, the MLN was reparametrized by a neural network evaluated on input features. A similar approach is the one in \cite{marra2019integrating}, where a continuous relaxation of the logical potentials allows for a faster inference in specific tasks. 
A related approach in the domain of logic programming is provided in \cite{manhaeve2018deepproblog}, where the probabilistic logic programming language ProbLog \citep{de2007problog} is extended to allow probabilities of atoms to be predicted by neural networks and to exploit differentiable algebraic generalizations of decision diagrams to train these networks. A common pattern in these approaches is that neural networks are "simply" exploited to parameterize a known relational model.
Compared to NMLN, they still rely on combinatorial structure learning (or rules hand-crafted by experts). Recently, there has been a renaissance of ILP methods in which neural computing is used to improve the search process  \citep{ellis:libraries,rocktaschel2017end, minervini2019differentiable,DBLP:journals/jair/SourekAZSK18}. These typically use templates or program sketches to reduce the size of the search space and gradient based methods to guide the search process. 
Unlike NMLN, none of these systems can model joint probability distributions over relational structures. 



\paragraph{NMLNs as KGE} NMLNs are also related to the many different knowledge graph embedding methods \citep{wang2017knowledge} as they can also exploit embeddings of domain elements and (implicitly) also relations. NMLNs and KGEs are most similar when $k=2$ and there are no unary and reflexive binary atoms. In this case, the NMLN still explicitly models the probabilistic relationship between different relations on the same pairs of constants, which KGE methods do not capture explicitly. Moreover, KGE methods cannot model unary relations (i.e.\ attributes) of the entities. Somewhat surprisingly, as noted by \citet{DBLP:conf/aaai/Kazemi018}, this problem is less well understood than link-prediction in KGEs. Furthermore, NMLNs can naturally incorporate both models of attributes and links, as demonstrated, e.g., on the small Smokers dataset and in the molecular generation experiment. Moreover, if we properly fix the neural architecture of NMLNs, many existing KGE methods can be explicitly modelled as NMLNs. So NMLNs are more flexible and can solve tasks that KGE methods cannot. On the other hand, KGEs are very fast and optimized for the KG completion tasks, as we also observed in our experiments (cf Section \ref{sec:exp:kge}).

\section{CONCLUSIONS}
\label{sec:conclusions}

We have introduced neural Markov logic networks, a statistical relational learning model combining representation learning power of neural networks with principled handling of uncertainty in the maximum-entropy framework. \mgedit{The proposed model works remarkably well both in small and large domains despite the fact that it actually solves a much harder problem (modelling joint probability distributions) than specialized models such as various knowledge graph embedding methods.}

\bibliography{biblio}
\bibliographystyle{unsrtnat}

\ifappendix

\appendix

\section{Gradient-based Optimization} 
\label{app:gradients}
The maximization of the log-likelihood is carried out by a gradient-based optimization scheme. The gradients of the log-likelihood w.r.t.\ both the parameters $w_{i,j}$, where $w_{i,j}$ denotes the $j$-th component of $\mathbf{w}_i$, and $\beta_i$ are:

\begin{align}
\label{eq:derivative_w}
\frac{\partial \log(P_{\widehat \omega})}{\partial w_{i,j}} & =  \beta_i  \bigg ( \frac{\partial \Phi_i(\widehat \omega; \mathbf{w}_i)}{\partial w_{i,j}}   -  \mathbb{E}_{ \omega \sim  P}\bigg[\frac{\partial \Phi(\omega;\mathbf{w}_i)}{\partial w_{i,j}}\bigg] \bigg )\\
\label{eq:derivative_beta}
\frac{\partial \log(P_{\widehat \omega})}{\partial \beta_i} & = \bigg ( \Phi_i(\widehat \omega; w_i)  -  \mathbb{E}_{ \omega \sim  P}[\Phi_i(\omega; w_i)] \bigg )
\end{align}

At a stationary point, Eq. \ref{eq:derivative_beta} recovers the initial constraint on statistics imposed in the maximization of the entropy. However, the minimization of the entropy is mapped to a new requirement: at stationary conditions, the expected value of the gradients of the $\Phi_i$ under the distribution must match the gradients of the $\Phi_i$ evaluated at the data points.

\section{Translating Markov Logic Networks to Neural Markov Logic Networks}
\label{app:mln}

In this section we show that any Markov logic network (MLN) without quantifiers can be represented as an NMLN.

\cite{kuvzelka2018relational} studies two maximum-entropy models, Model A, which is close to the model that we study in this paper, and Model B, which is the same as MLNs. Syntactically, both models are encoded as sets of quantifier-free weighted first order logic formulas, e.g.\ $\Phi = \{ (\alpha_1,w_1), \dots, (\alpha_M,w_m)\}$. In particular, given a positive integer $k$, Model A defines the following distribution:
$$p_A(\omega) = \frac{1}{Z} \exp{\left( \sum_{(\alpha,w) \in \Phi_A} w \cdot \#_k(\alpha,\omega) \right)}$$
where $Z$ is the normalization constant and $\#(\alpha,\omega)$ is the fraction of size-$k$ subsets $\mathcal{S}$ of constants in the possible world $\omega$ for which $\omega\langle \mathcal{S} \rangle \models \alpha$ (i.e. the formula $\alpha$ is classically true in the fragment of $\omega$ induced by $\mathcal{S}$). 
Let us first define
$$\phi_{\alpha,w}(\gamma) = \begin{cases} w & \gamma \models \alpha \\ 0 & \gamma \not\models \alpha \end{cases}$$
It is then easy to see that the distribution $p_A(\omega)$ can also easily be encoded as an NMLN by selecting the potential function $\phi(\gamma) = \sum_{(\alpha,w) \in \Phi_A} \phi_{\alpha,w}(\gamma)$ 
and by carefully selecting the weights $\beta_i$ in the NMLN.

Next we show that all distributions in Model B can be translated to distributions in Model A. First we will assume that the formulas $\alpha_i$ do not contain any constants.

Model B is given by
$$p_B(\omega) = \frac{1}{Z} \exp{\left( \sum_{(\beta,v) \in \Phi_B} v \cdot n(\beta,\omega) \right)}$$
where $n(\beta,\omega)$ is the number\footnote{In \citep{kuvzelka2018relational}, Model B is defined using {\it fractions} of true grounding substitutions instead of {\it numbers} of true grounding substitutions. However, these two definitions are equivalent up to normalizations and both work for our purposes but the latter one is a bit more convenient here. Hence we choose the latter one here.} of true injective groundings of the formula $\beta$ in the possible world $\omega$. Hence, Model B is exactly the same as Markov logic networks up to the requirement on injectivity of the groundings. However, as shown in \citep{buchman2015representing}, any Markov logic network can be translated into such modified Markov logic network with the injectivity requirement on the groundings. 

Let $k$ be an integer greater or equal to the number of variables in any formula in $\Phi_B$. Now, let $\Gamma$ be the set of all size-$k$ fragments. For every formula $\beta$ in $\Phi_B$, we introduce a partition $\mathcal{P}$ on $\Gamma$ induced by the equivalence relation $\sim_\beta$ defined by: $\gamma \sim_\beta \gamma'$ iff $n(\beta,\gamma) = n(\beta,\gamma')$. Since $\beta$ is assumed to not contain any constants, we can capture each of these equivalence classes $C$ by a (possibly quite big) first-order logic sentence without constants $\beta_C$. Let $C_i$ be the equivalence class that contains fragments $\gamma$ such that $n(\beta,\gamma)=i$. Let $m(\beta,\omega) = \sum_{C_i \in \mathcal{P}}\sum_{\gamma \in \Gamma_k(\omega)} i \cdot \mathds{1}(\gamma \models \beta_C)$. By construction, it holds $m(\beta,\omega) = \sum_{\gamma \in \Gamma_k(\omega)} n(\beta,\gamma)$. Every true injective grounding of the formula $\beta$, having $l$ variables, is contained in $\binom{n-l}{k-l}$ different size-$k$ fragments of $\omega$, each of which gives rise to $k!$ anonymized fragments in the multi-set $\Gamma_k(\omega)$. So $m(\beta,\omega)$ is over-counting the number of true groundings $n(\beta,\omega)$ by a constant factor. It follows that, by carefully selecting the weights of the formulas $\beta_C$ we can encode the distribution $p_B(\omega)$ also in Model A. Although this particular transformation that we have just sketched is not very efficient, it does show that NMLNs with potential functions of width $k$ can express all distributions that can be expressed by MLNs containing formulas with at most $k$ variables and no existential quantifiers.

First-order logic formulas defining MLNs may also contain constants. In NMLNs we may represent constants using vector-space embeddings as described in the main text. One can then easily extend the argument sketched above to the case covering MLNs with constants.

\section{Generating Molecules}
\label{app:molecules}

\subsection{Molecules First-Order-logic representation}
Even though molecules can be described with a high level of precision, using both spatial features (i.e. atoms distances, bond length etc.) and chemical features (i.e. atom charge, atom mass, hybridization), in this work, we focused only on structural symbolic descriptions of molecules.

In particular, we described a molecule using three sets of FOL predicates:
\begin{itemize}
    \item \textit{Atom-type unary predicates}: these are \texttt{C}, \texttt{N}, \texttt{O}, \texttt{S}, \texttt{Cl}, \texttt{F}, \texttt{P}.
    \item \textit{Bond-type binary predicate}: these are \texttt{SINGLE} and \texttt{DOUBLE}.
    \item an auxiliary binary predicate \texttt{SKIPBOND} (see later).
\end{itemize}

An example of a molecule FOL description can be:

{\small
{\tt O(0)}, {\tt C(1)}, {\tt C(2)}, {\tt C(3)}, {\tt N(4)}, {\tt C(5)}, {\tt C(6)}, {\tt C(7)}, {\tt O(8)}, {\tt O(9),}
{\tt SINGLE(0,1)}, {\tt SINGLE(1,0)}, {\tt SINGLE(1,2)}, {\tt SINGLE(2,1)}, {\tt SINGLE(2,3),} 
{\tt SINGLE(3,2)}, {\tt SINGLE(3,4)}, {\tt SINGLE(4,3)}, {\tt SINGLE(4,5)}, {\tt SINGLE(5,4),}
{\tt SINGLE(5,6)}, {\tt SINGLE(6,5)}, {\tt SINGLE(5,7)}, {\tt SINGLE(7,5)}, {\tt DOUBLE(7,8),}
{\tt DOUBLE(8,7)}, {\tt SINGLE(7,9)}, {\tt SINGLE(9,7)}, {\tt SINGLE(6,1)}, {\tt SINGLE(1,6)}
}

To help NMLNs capture long-range dependencies in molecules even with not so large fragments (e.g. $k=3,4,5$), we added ``skip-bond'' atoms. For any three distinct $x$, $y$, $z$, such that there is a bond connecting $x$ and $y$ and a bond connecting $y$ and $z$, we add {\tt SKIPBOND(x,z)}. This forces NMLN to learn the ``definition'' of skip-bonds and allows them, for instance, to implicitly capture the presence of a six-ring of atoms with a fragment of size $4$ (whereas without skip-bonds we would need fragments of size $6$ for that).

\subsection{Faster inference with chemical constraints.}
To speed up the convergence, we exploited a slightly modified Gibbs Sampling procedure, again inspired by the blocking technique. In particular, given a constant or a pair of constants, we sample all its relations in parallel. Since we known, that a constant should belong to \textit{exactly one} unary relation, i.e.\ atom type, and a pair of constants should belong to \textit{at most one} bond type, we can shorten the length of the chain by avoiding sampling inconsistent states. Inconsistent states are simply defined in NMLN as a set of constraints over the chain intermediate states. We also executed the same generation experiment without any constraint and there were no sensible differences. 

\begin{figure}
\centering
\includegraphics[trim=0.5cm 11cm 0.5cm 10cm,width=0.7\linewidth]{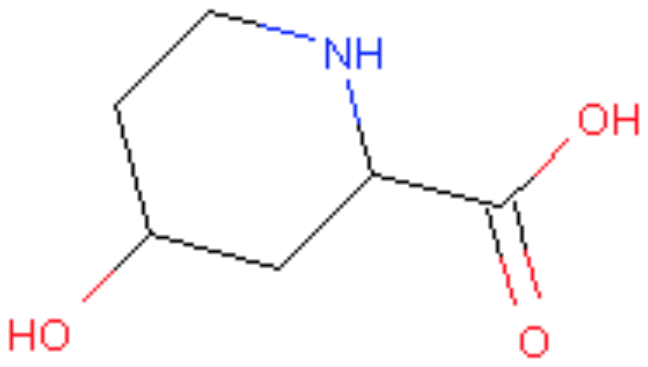}
\caption{An example of molecule}
\label{fig:moleculeexample}
\end{figure}

We show a random sample of the training data (Figure \ref{fig:true_mols}) and the most frequent molecules sampled by GS (Figure \ref{fig:gen_mols}).


\subsection{Training, Generalization and Novelty}

In this Section, we describe the training and generalization  capabilities of NMLNs in the molecule generation experiment. 

First of all, we measured the number of generated molecules which belongs to the training set. We wanted to use this knowledge as an indication of the generalization capability of the model in terms of the novelty (w.r.t. the training set) of the generated molecules.
The results are shown in Figure~\ref{fig:molecules_training_support}. It is shown that 53 out of the 100 most frequently generated molecules are indeed in the training set, which is a good sign, because we expect training molecules to be very likely. However, for larger number of generated molecules, NMLNs generate lots of new molecules, that have never been observed during training process.

\begin{figure}
    \centering
    \includegraphics[width=0.6\linewidth]{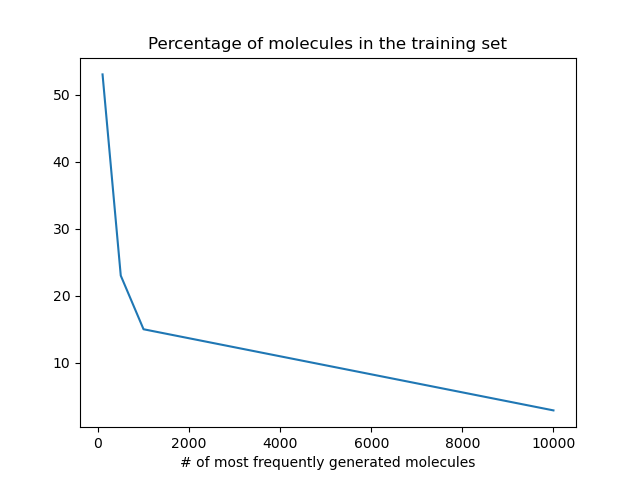}
    \caption{\textbf{Molecules generation.} Percentage of generated molecules in the training set when increasing the number of generated molecules.}
    \label{fig:molecules_training_support}
\end{figure}
However, these numbers still do not tell us whether the novel generated molecules (i.e. the ones not in the trainig set) are meaningful or not. Since we reject generated chemically invalid molecules during training (since they can be just checked with an automatic tool like RDkit, we get a method reminiscent of rejection sampling), all the generated molecules are chemically valid and thus we check weather they are known in larger databases than the one we used for training. We selected 100\footnote{These chemical online databases limit the number of queries that one can do for free.} of the novel generated molecules and we checked in the ChemSpider dataset\footnote{\url{http://www.chemspider.com}} if they are indexed or not. \textbf{83} out of 100 molecules are indeed existing molecules, which are shown in Table~\ref{tab:molecules_novel}. We still don't know if the remaining 17 molecules are simply not indexed in ChemSpider, are impossible for more complex chemical reasons (not checkable with RDkit) or they just represent completely novel molecules. However, it is rather impressive that most of these molecules that NMLNs generated are molecules that actually exist and were not present in the training data.

\section{Implementation Details}
\label{app:impl}

We provide details about hyperparameters exploited in the experiments described in the main text.

All the functions $\phi$ were implemented using feed-forward neural networks. No  regularization technique is exploited on the weights or activations of the network (e.g. L2, dropout), even though, as highlighted in the main text, the addition of noise has a regularizing effect apart from avoiding NMLN to focus on deterministic rules.

When we searched over grids (grid-search), we list all the values and we show in bold the selected ones.

\paragraph{NMLN  (Smokers)}
\begin{itemize}
    \item Network architecture: 1 hidden layer with 30 sigmoidal neurons. 
    \item Number of parallel chains: 10
    \item $\pi_n$: 0.
    \item learning rates: [0.1, \textbf{0.01}, 0.001]
\end{itemize}

\paragraph{NMLN-K3 (Nations)}
\begin{itemize}
    \item Network architecture: 2 hidden layer with [150,\textbf{100},75] and [\textbf{50},25] [\textbf{ReLU},sigmoid] neurons. 
    \item Number of parallel chains: 10
    \item $\pi_n$: [\textbf{0.01},0.02,0.03]
    \item learning rates: [$10^{-4}$,$\pmb{10^{-5}}$,$10^{-6}$]
\end{itemize}

\paragraph{NMLN-K3 (Kinship, UMLS)}

We just run them on the best performing configuration on Nations.

\paragraph{NMLN-K2E (Nations, Kinship, UMLS)}
\begin{itemize}
	\item Network architecture: 2 hidden layer with 75 and 50 ReLU neurons. 
	\item Number of parallel chains: 10
	\item $\pi_n$: [0.01,0.02,0.03]
	\item learning rates: [0.1, 0.01, 0.001]
	\item Embedding size: [10,20,30]
	\item Number of disconnected fragments per connected one: 2
\end{itemize}
For $\pi_n$, learning rate and embedding size the selected configurations are, respectively for each datasets: Nations (\textbf{0.02, 0.001, 20}), Kinship (\textbf{0.02, 0.1, 10}), UMLS (\textbf{0.02, 0.01, 30})

\paragraph{NMLN-K2E (WordNet, Freebase)}
\begin{itemize}
	\item Network architecture: 3 hidden layer with 50,50 and 50 ReLU neurons. 
	\item Number of parallel chains: 10
	\item $\pi_n$: [0.01,0.03, 0.04, 0.05, 0.1]
	\item learning rates: [0.1, 0.01, 0.001]
	\item Embedding size: [5,10,30]
	\item Number of disconnected fragments per connected one: [2,4,10]
\end{itemize}
For $\pi_n$, learning rate, embedding size and the number of disconnected fragments per connected one, the selected configurations are, respectively for each datasets: Wordnet (\textbf{0.05, 0.01, 10, 2}), Freebase (\textbf{0.04, 0.01, 10, 2})

\paragraph{Molecules generation}

\begin{itemize}
	\item Network architecture: 2 hidden layer with 150 and 50 ReLU neurons. 
	\item Number of parallel chains: \textbf{10}
	\item $\pi_n$: [0., \textbf{0.01}]
	\item learning rates: [\textbf{0.0001}]
\end{itemize}

\comm{This can be easily removed to obtain more space.}
\comm{When exploiting the subsampling of disconnected fragments in \textit{NMLN-T}, we are approximating the correct training principle described in the paper. This approximation requires two implementation choices to be made: \textit{(i)} the number $d$ of disconnected fragments per connected one, and \textit{(ii)} the way we aggregate connected and disconnected fragments in the potential computation. Theoretically, one would say that \textit{(i)} larger $d$s recover less approximated solutions, and, \textit{(ii)} we should weight more the disconnected fragments, in order to rebalance the subsampling. While these choices are principled in order to obtain more precise marginal likelihoods, it is not necessarily the case for ranking metrics, like $HITS\text{@}m$ or $MRR$. Empirically, we observed that better performances always come with small $d$s and unweighted aggregations. One could attribute this behaviour to the fact that simpler choices are more robust to the noise introduced by the subsampling, but we do not investigated further in this direction and we leave this for future works.}

\onecolumn
\begin{figure*}
  \centering
  \includegraphics[width=0.8\linewidth]{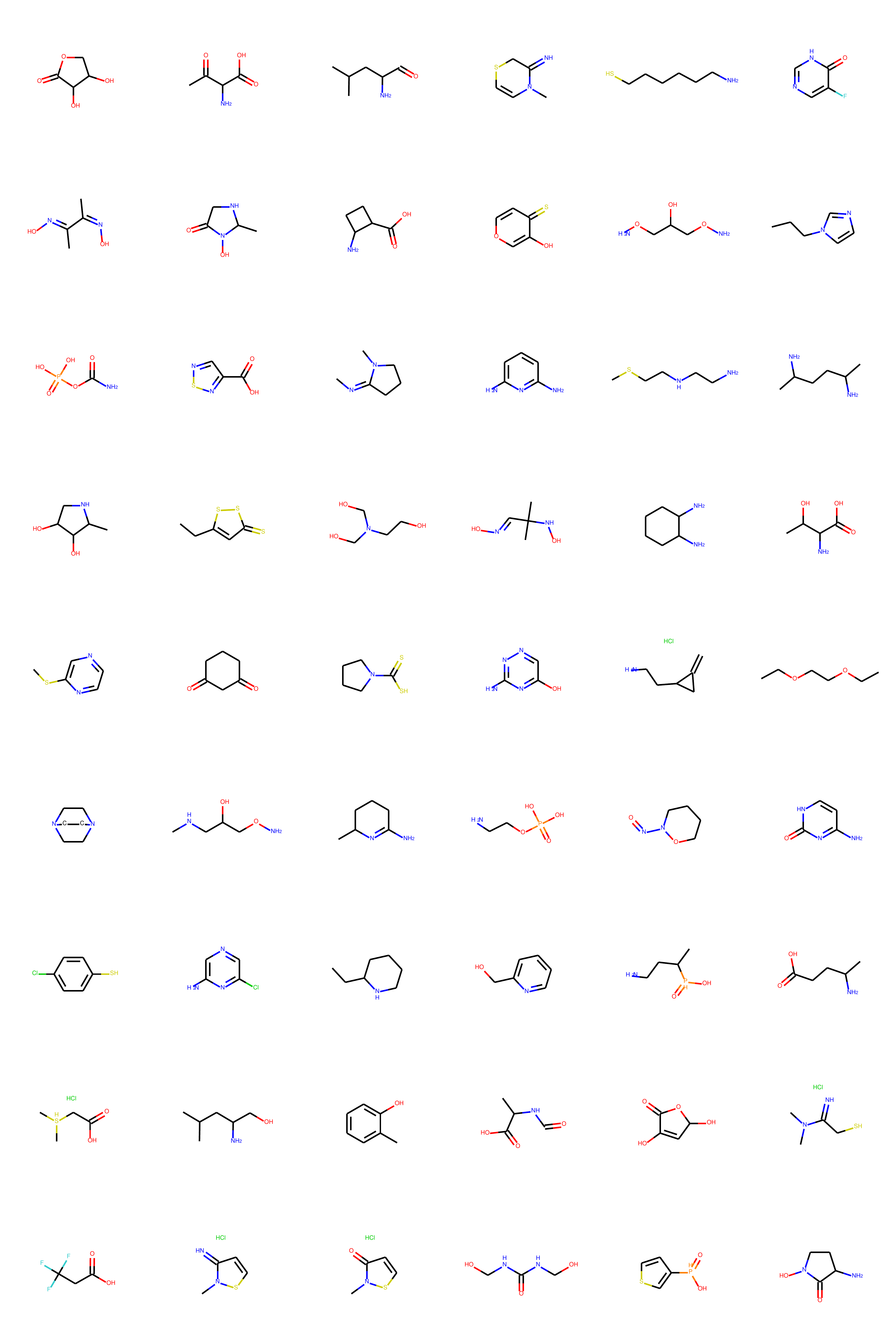}
  \caption{Molecules from the training data.}
  \label{fig:true_mols}
\end{figure*}

\begin{figure*}  \centering
  \includegraphics[width=0.8\linewidth]{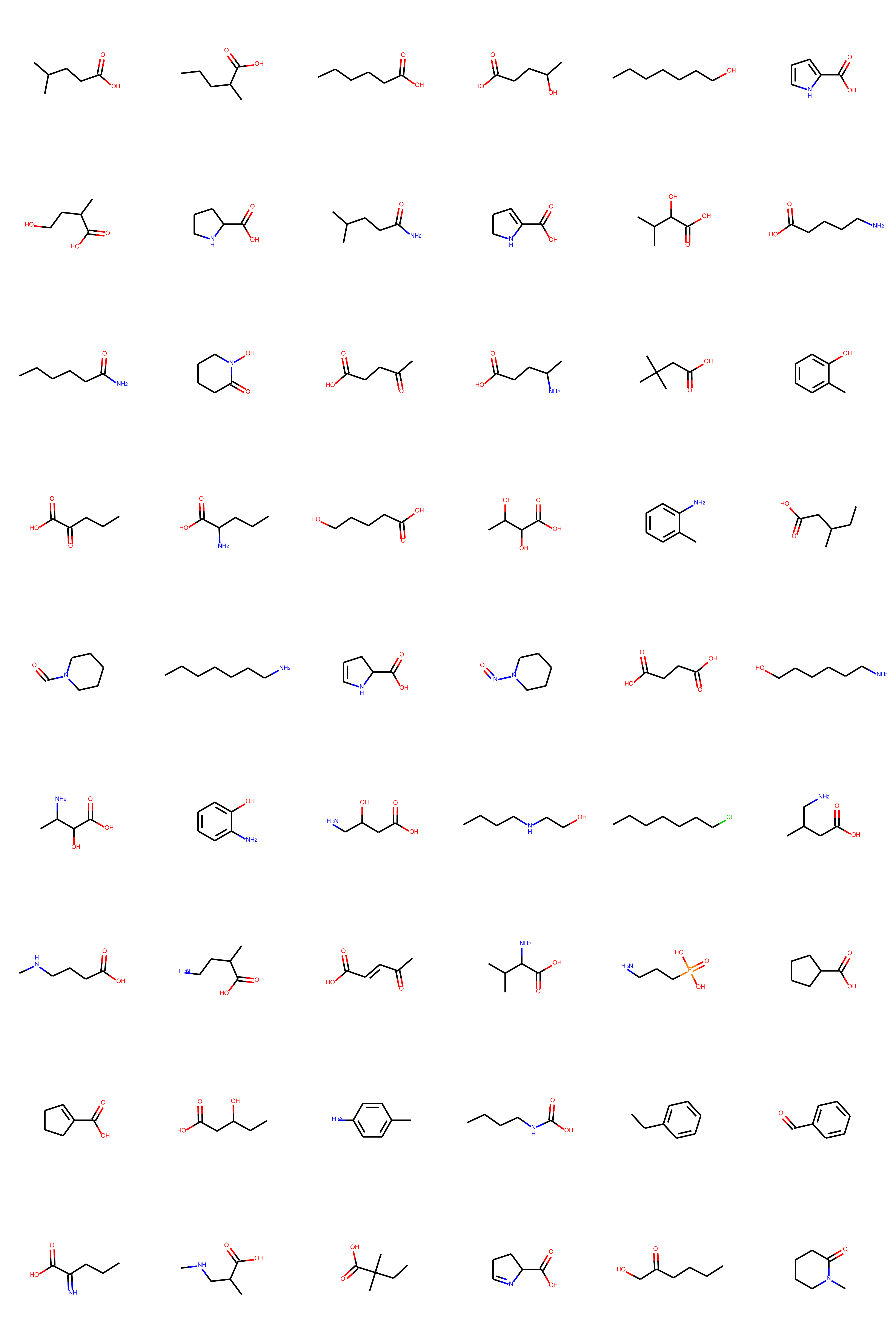}
  \caption{Most frequent generated molecules.}
  \label{fig:gen_mols}
\end{figure*}

\begin{longtable}{|c|c|c|}
\caption{Novel Molecules existing in ChemSpider}
\label{tab:molecules_novel} \\ 
\hline
\textbf{Smile code} & \textbf{Molecule} & \textbf{Name}\\ 
\hline
\endfirsthead
\multicolumn{3}{c}%
{\tablename\ \thetable\ -- \textit{Continued from previous page}} \\ 
\hline
\textbf{Smile code} & \textbf{Molecule} & \textbf{Name} \\
\hline
\endhead

\begin{tabular}{l}CC(C)CCC(N)=O\end{tabular} &  \begin{tabular}{l}\includegraphics[width=0.1\linewidth]{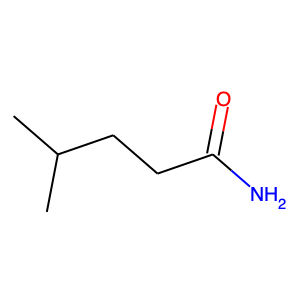}\end{tabular}& \begin{tabular}{l} 4-Methylpentanamide\end{tabular} \\ \hline 
\begin{tabular}{l}O=C(O)C1=CCCN1\end{tabular} &  \begin{tabular}{l}\includegraphics[width=0.1\linewidth]{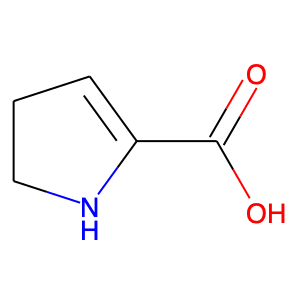}\end{tabular}& \begin{tabular}{l}Multiple search results\end{tabular} \\ \hline 
\begin{tabular}{l}O=C1CCCCN1O\end{tabular} &  \begin{tabular}{l}\includegraphics[width=0.1\linewidth]{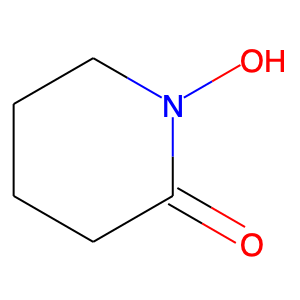}\end{tabular}& \begin{tabular}{l} 1-Hydroxy-2-piperidinone\end{tabular} \\ \hline 
\begin{tabular}{l}CC(N)CCC(=O)O\end{tabular} &  \begin{tabular}{l}\includegraphics[width=0.1\linewidth]{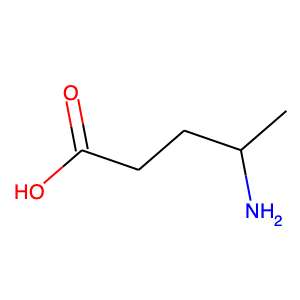}\end{tabular}& \begin{tabular}{l} 4-AMINOVALERIC ACID\end{tabular} \\ \hline 
\begin{tabular}{l}CCCC(N)C(=O)O\end{tabular} &  \begin{tabular}{l}\includegraphics[width=0.1\linewidth]{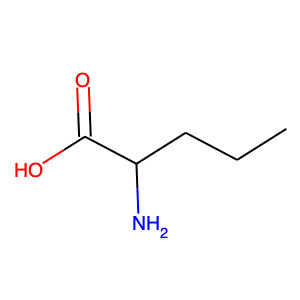}\end{tabular}& \begin{tabular}{l} DL-Norvaline\end{tabular} \\ \hline 
\begin{tabular}{l}CC(O)C(O)C(=O)O\end{tabular} &  \begin{tabular}{l}\includegraphics[width=0.1\linewidth]{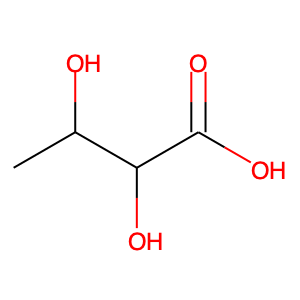}\end{tabular}& \begin{tabular}{l} 4-DEOXYTETRONIC ACID\end{tabular} \\ \hline 
\begin{tabular}{l}NCCCCCCO\end{tabular} &  \begin{tabular}{l}\includegraphics[width=0.1\linewidth]{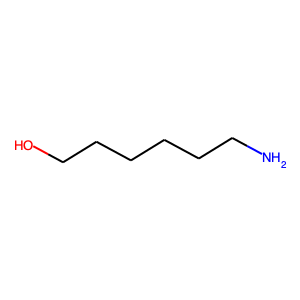}\end{tabular}& \begin{tabular}{l} MO8840000\end{tabular} \\ \hline 
\begin{tabular}{l}CC(N)C(O)C(=O)O\end{tabular} &  \begin{tabular}{l}\includegraphics[width=0.1\linewidth]{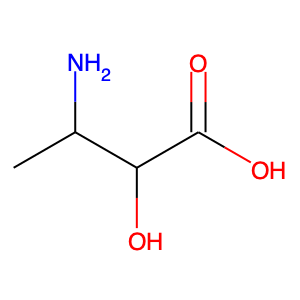}\end{tabular}& \begin{tabular}{l} 3-amino-2-hydroxybutanoic acid\end{tabular} \\ \hline 
\begin{tabular}{l}CC(CCN)C(=O)O\end{tabular} &  \begin{tabular}{l}\includegraphics[width=0.1\linewidth]{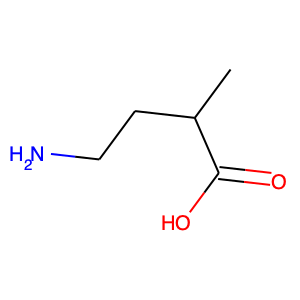}\end{tabular}& \begin{tabular}{l} 4-Amino-2-methylbutanoic acid\end{tabular} \\ \hline 
\begin{tabular}{l}CC(=O)C=CC(=O)O\end{tabular} &  \begin{tabular}{l}\includegraphics[width=0.1\linewidth]{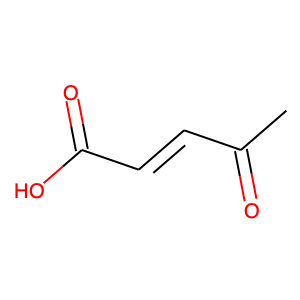}\end{tabular}& \begin{tabular}{l} 4-Oxo-2-pentenoic acid\end{tabular} \\ \hline 
\begin{tabular}{l}CCC(O)CC(=O)O\end{tabular} &  \begin{tabular}{l}\includegraphics[width=0.1\linewidth]{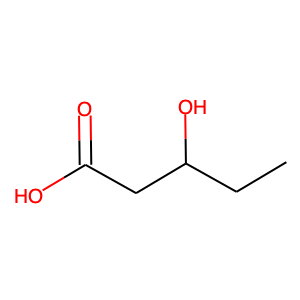}\end{tabular}& \begin{tabular}{l} 3-Hydroxyvaleric acid\end{tabular} \\ \hline 
\begin{tabular}{l}CCCCNC(=O)O\end{tabular} &  \begin{tabular}{l}\includegraphics[width=0.1\linewidth]{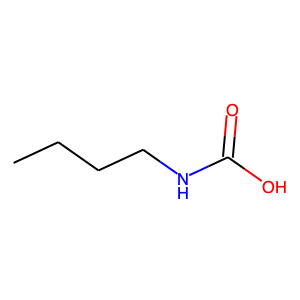}\end{tabular}& \begin{tabular}{l} Butylcarbamic acid\end{tabular} \\ \hline 
\begin{tabular}{l}CCCC(=N)C(=O)O\end{tabular} &  \begin{tabular}{l}\includegraphics[width=0.1\linewidth]{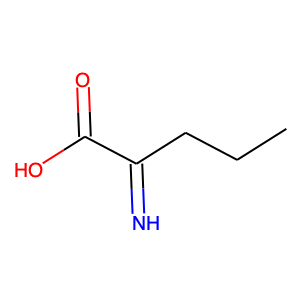}\end{tabular}& \begin{tabular}{l}Multiple search results\end{tabular} \\ \hline 
\begin{tabular}{l}CCCCC(=O)CO\end{tabular} &  \begin{tabular}{l}\includegraphics[width=0.1\linewidth]{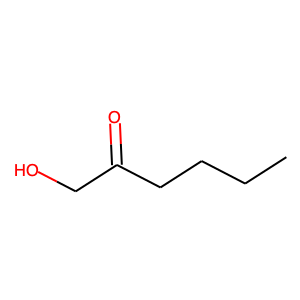}\end{tabular}& \begin{tabular}{l} 1-Hydroxy-2-hexanone\end{tabular} \\ \hline 
\begin{tabular}{l}OCCCNCCO\end{tabular} &  \begin{tabular}{l}\includegraphics[width=0.1\linewidth]{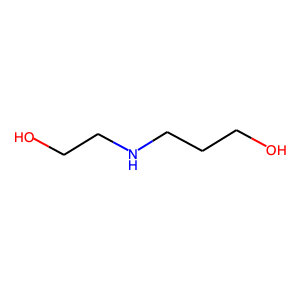}\end{tabular}& \begin{tabular}{l} 3-((2-Hydroxyethyl)amino)propanol\end{tabular} \\ \hline 
\begin{tabular}{l}NC(=O)CCC(=O)O\end{tabular} &  \begin{tabular}{l}\includegraphics[width=0.1\linewidth]{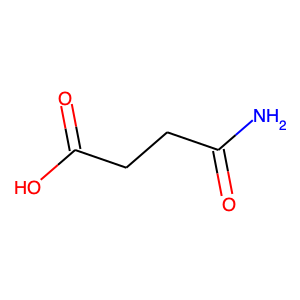}\end{tabular}& \begin{tabular}{l} 4-Amino-4-oxobutanoic acid\end{tabular} \\ \hline 
\begin{tabular}{l}O=C(O)C1C=CCC1\end{tabular} &  \begin{tabular}{l}\includegraphics[width=0.1\linewidth]{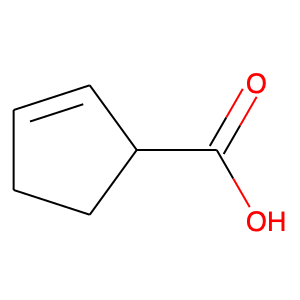}\end{tabular}& \begin{tabular}{l} 2-Cyclopentenecarboxylic acid\end{tabular} \\ \hline 
\begin{tabular}{l}ONC1CCCCC1\end{tabular} &  \begin{tabular}{l}\includegraphics[width=0.1\linewidth]{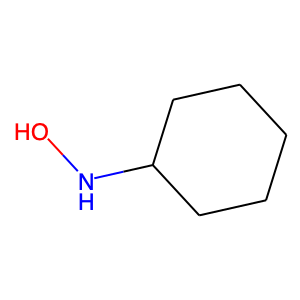}\end{tabular}& \begin{tabular}{l} NC3410400\end{tabular} \\ \hline 
\begin{tabular}{l}CC(C)CCCCO\end{tabular} &  \begin{tabular}{l}\includegraphics[width=0.1\linewidth]{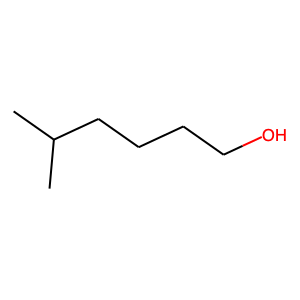}\end{tabular}& \begin{tabular}{l} 5-Methyl-1-hexanol\end{tabular} \\ \hline 
\begin{tabular}{l}CC(=O)CCC(N)=O\end{tabular} &  \begin{tabular}{l}\includegraphics[width=0.1\linewidth]{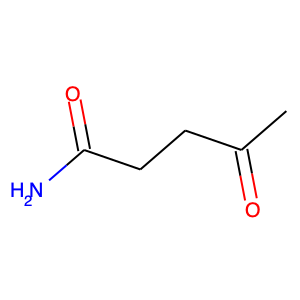}\end{tabular}& \begin{tabular}{l} 4-Oxopentanamide\end{tabular} \\ \hline 
\begin{tabular}{l}NC1CCCC(=O)C1\end{tabular} &  \begin{tabular}{l}\includegraphics[width=0.1\linewidth]{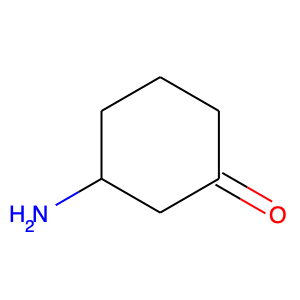}\end{tabular}& \begin{tabular}{l} 3-Aminocyclohexanone\end{tabular} \\ \hline 
\begin{tabular}{l}O=NC1CCCCC1\end{tabular} &  \begin{tabular}{l}\includegraphics[width=0.1\linewidth]{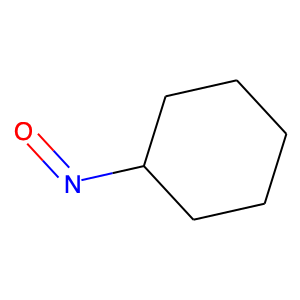}\end{tabular}& \begin{tabular}{l} Nitrosocyclohexane\end{tabular} \\ \hline 
\begin{tabular}{l}CCCCP(=O)(O)O\end{tabular} &  \begin{tabular}{l}\includegraphics[width=0.1\linewidth]{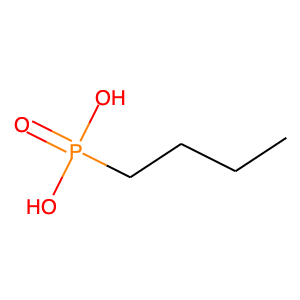}\end{tabular}& \begin{tabular}{l} Butylphosphonate\end{tabular} \\ \hline 
\begin{tabular}{l}CC(C)COC(=O)O\end{tabular} &  \begin{tabular}{l}\includegraphics[width=0.1\linewidth]{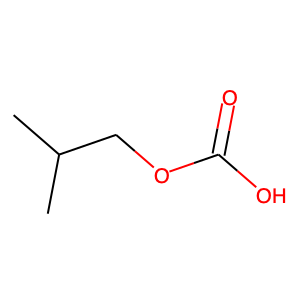}\end{tabular}& \begin{tabular}{l} Isobutyl hydrogen carbonate\end{tabular} \\ \hline 
\begin{tabular}{l}CCCCCNC=O\end{tabular} &  \begin{tabular}{l}\includegraphics[width=0.1\linewidth]{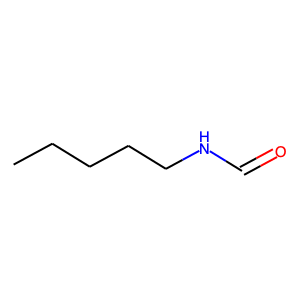}\end{tabular}& \begin{tabular}{l} MFCD07784338\end{tabular} \\ \hline 
\begin{tabular}{l}CC(C)(O)CC(=O)O\end{tabular} &  \begin{tabular}{l}\includegraphics[width=0.1\linewidth]{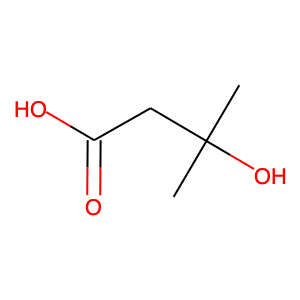}\end{tabular}& \begin{tabular}{l} 3-OH-isovaleric acid\end{tabular} \\ \hline 
\begin{tabular}{l}O=C1CCCCC1O\end{tabular} &  \begin{tabular}{l}\includegraphics[width=0.1\linewidth]{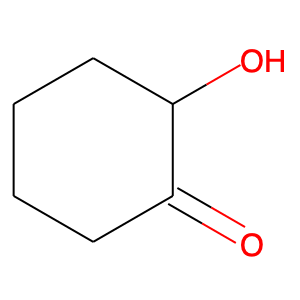}\end{tabular}& \begin{tabular}{l} Adipoin\end{tabular} \\ \hline 
\begin{tabular}{l}CNCCCCCO\end{tabular} &  \begin{tabular}{l}\includegraphics[width=0.1\linewidth]{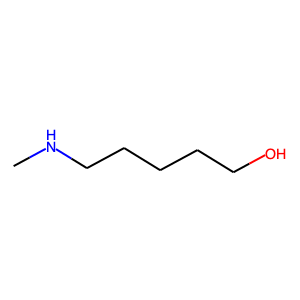}\end{tabular}& \begin{tabular}{l} MFCD16696454\end{tabular} \\ \hline 
\begin{tabular}{l}CCCC(C)C(C)O\end{tabular} &  \begin{tabular}{l}\includegraphics[width=0.1\linewidth]{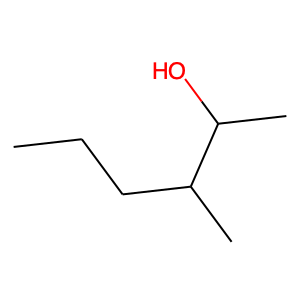}\end{tabular}& \begin{tabular}{l} MFCD00021889\end{tabular} \\ \hline 
\begin{tabular}{l}CC(C)CCC(N)=S\end{tabular} &  \begin{tabular}{l}\includegraphics[width=0.1\linewidth]{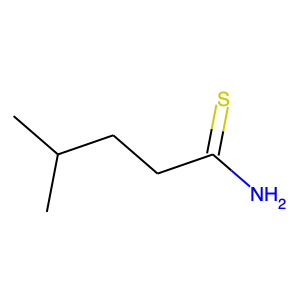}\end{tabular}& \begin{tabular}{l} 4-Methylpentanethioamide\end{tabular} \\ \hline 
\begin{tabular}{l}O=C(O)C1C=CC=N1\end{tabular} &  \begin{tabular}{l}\includegraphics[width=0.1\linewidth]{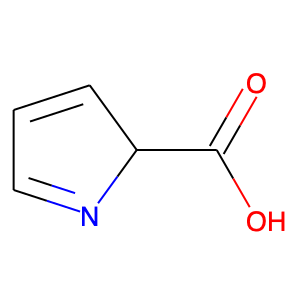}\end{tabular}& \begin{tabular}{l}Multiple search results\end{tabular} \\ \hline 
\begin{tabular}{l}CCCCCC(C)C\end{tabular} &  \begin{tabular}{l}\includegraphics[width=0.1\linewidth]{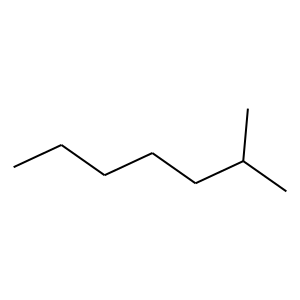}\end{tabular}& \begin{tabular}{l} 2-Methylheptane\end{tabular} \\ \hline 
\begin{tabular}{l}CC(C)C=CC(N)=O\end{tabular} &  \begin{tabular}{l}\includegraphics[width=0.1\linewidth]{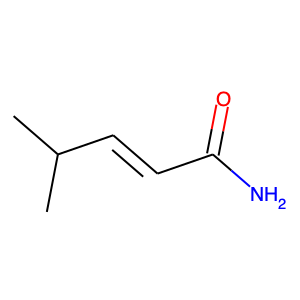}\end{tabular}& \begin{tabular}{l} 4-Methyl-2-pentenamide\end{tabular} \\ \hline 
\begin{tabular}{l}COCCCC(=O)O\end{tabular} &  \begin{tabular}{l}\includegraphics[width=0.1\linewidth]{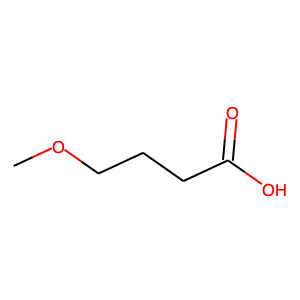}\end{tabular}& \begin{tabular}{l} 4-Methoxybutanoic acid\end{tabular} \\ \hline 
\begin{tabular}{l}NCCC(=O)C(=O)O\end{tabular} &  \begin{tabular}{l}\includegraphics[width=0.1\linewidth]{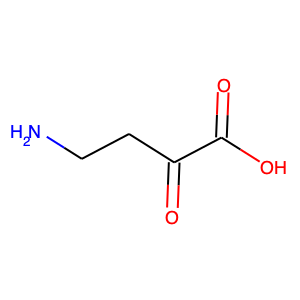}\end{tabular}& \begin{tabular}{l} 4-Amino-2-oxobutanoic acid\end{tabular} \\ \hline 
\begin{tabular}{l}NCC(=O)CC(=O)O\end{tabular} &  \begin{tabular}{l}\includegraphics[width=0.1\linewidth]{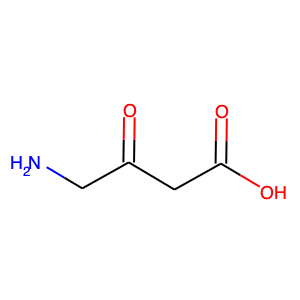}\end{tabular}& \begin{tabular}{l} 4-Amino-3-oxobutanoic acid\end{tabular} \\ \hline 
\begin{tabular}{l}CCCC=CC(=O)O\end{tabular} &  \begin{tabular}{l}\includegraphics[width=0.1\linewidth]{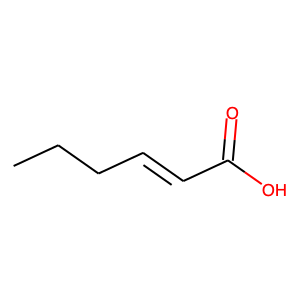}\end{tabular}& \begin{tabular}{l} 2-Hexenoic acid\end{tabular} \\ \hline 
\begin{tabular}{l}CCCCCC(O)O\end{tabular} &  \begin{tabular}{l}\includegraphics[width=0.1\linewidth]{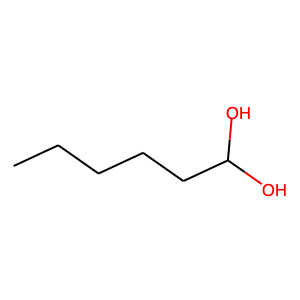}\end{tabular}& \begin{tabular}{l} Hexanediol\end{tabular} \\ \hline 
\begin{tabular}{l}CCCC(=O)C(C)O\end{tabular} &  \begin{tabular}{l}\includegraphics[width=0.1\linewidth]{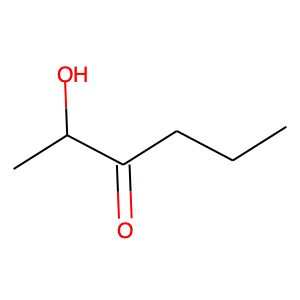}\end{tabular}& \begin{tabular}{l} 2-Hydroxy-3-hexanone\end{tabular} \\ \hline 
\begin{tabular}{l}CC(CO)CC(N)=O\end{tabular} &  \begin{tabular}{l}\includegraphics[width=0.1\linewidth]{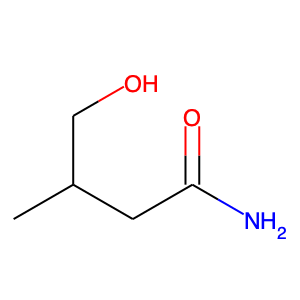}\end{tabular}& \begin{tabular}{l} 4-Hydroxy-3-methylbutanamide\end{tabular} \\ \hline 
\begin{tabular}{l}CCCCCC(=N)N\end{tabular} &  \begin{tabular}{l}\includegraphics[width=0.1\linewidth]{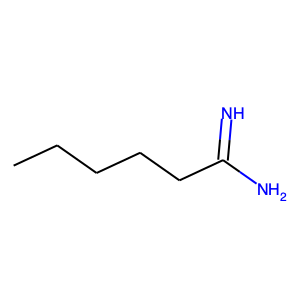}\end{tabular}& \begin{tabular}{l} Hexanamidine\end{tabular} \\ \hline 
\begin{tabular}{l}CC1CCC1C(=O)O\end{tabular} &  \begin{tabular}{l}\includegraphics[width=0.1\linewidth]{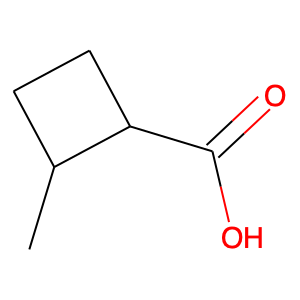}\end{tabular}& \begin{tabular}{l} 2-Methylcyclobutanecarboxylic acid\end{tabular} \\ \hline 
\begin{tabular}{l}NC(CCO)C(=O)O\end{tabular} &  \begin{tabular}{l}\includegraphics[width=0.1\linewidth]{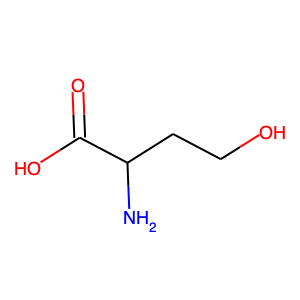}\end{tabular}& \begin{tabular}{l} Homoserine\end{tabular} \\ \hline 
\begin{tabular}{l}O=C1CC=CC=C1O\end{tabular} &  \begin{tabular}{l}\includegraphics[width=0.1\linewidth]{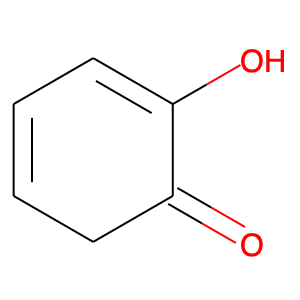}\end{tabular}& \begin{tabular}{l}Multiple search results\end{tabular} \\ \hline 
\begin{tabular}{l}CC(C)CCCCN\end{tabular} &  \begin{tabular}{l}\includegraphics[width=0.1\linewidth]{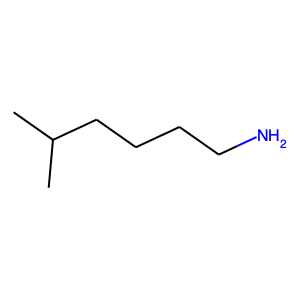}\end{tabular}& \begin{tabular}{l} 5-Methyl-1-hexanamine\end{tabular} \\ \hline 
\begin{tabular}{l}NC1=CC=CC(=O)C1\end{tabular} &  \begin{tabular}{l}\includegraphics[width=0.1\linewidth]{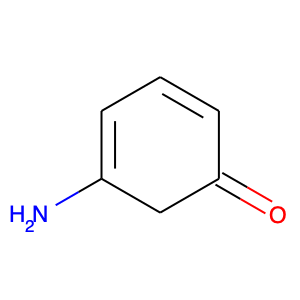}\end{tabular}& \begin{tabular}{l}Multiple search results\end{tabular} \\ \hline 
\begin{tabular}{l}CNC1CCCCC1\end{tabular} &  \begin{tabular}{l}\includegraphics[width=0.1\linewidth]{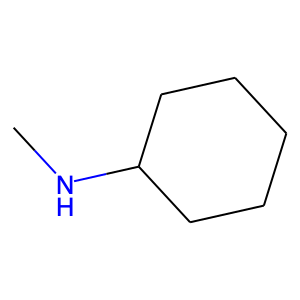}\end{tabular}& \begin{tabular}{l} GX1529000\end{tabular} \\ \hline 
\begin{tabular}{l}CCCC1=CC(=O)C1\end{tabular} &  \begin{tabular}{l}\includegraphics[width=0.1\linewidth]{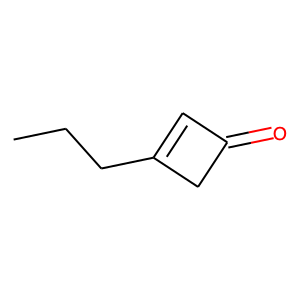}\end{tabular}& \begin{tabular}{l} 3-Propyl-2-cyclobuten-1-one\end{tabular} \\ \hline 
\begin{tabular}{l}ClC1=CCCCCC1\end{tabular} &  \begin{tabular}{l}\includegraphics[width=0.1\linewidth]{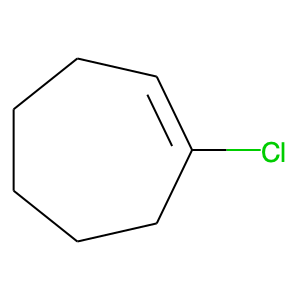}\end{tabular}& \begin{tabular}{l} 1-Chlorocycloheptene\end{tabular} \\ \hline 
\begin{tabular}{l}CCCCOC(=O)O\end{tabular} &  \begin{tabular}{l}\includegraphics[width=0.1\linewidth]{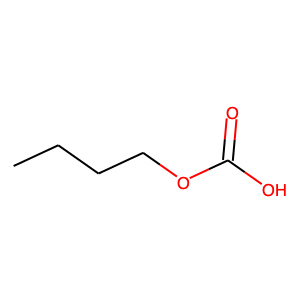}\end{tabular}& \begin{tabular}{l} monobutyl carbonate\end{tabular} \\ \hline 
\begin{tabular}{l}O=C(Cl)CC1C=CC1\end{tabular} &  \begin{tabular}{l}\includegraphics[width=0.1\linewidth]{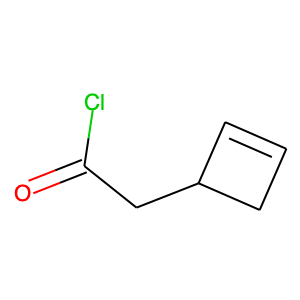}\end{tabular}& \begin{tabular}{l} Cyclobutylideneacetyl chloride\end{tabular} \\ \hline 
\begin{tabular}{l}CC(=O)CCCCN\end{tabular} &  \begin{tabular}{l}\includegraphics[width=0.1\linewidth]{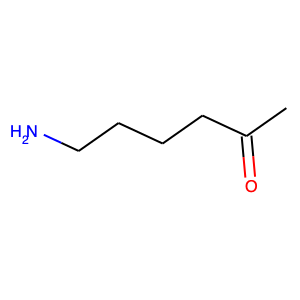}\end{tabular}& \begin{tabular}{l} 6-Amino-2-hexanone\end{tabular} \\ \hline 
\begin{tabular}{l}CCNC(C)C(=O)O\end{tabular} &  \begin{tabular}{l}\includegraphics[width=0.1\linewidth]{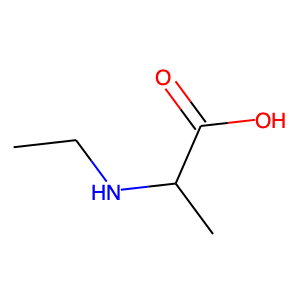}\end{tabular}& \begin{tabular}{l} N-Ethylalanine\end{tabular} \\ \hline 
\begin{tabular}{l}N=C1CCCCN1O\end{tabular} &  \begin{tabular}{l}\includegraphics[width=0.1\linewidth]{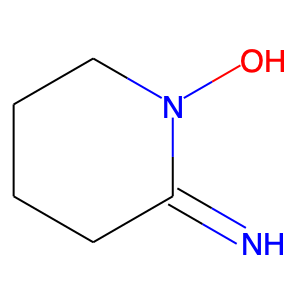}\end{tabular}& \begin{tabular}{l} 2-Imino-1-piperidinol\end{tabular} \\ \hline 
\begin{tabular}{l}CCCC=CC(N)=O\end{tabular} &  \begin{tabular}{l}\includegraphics[width=0.1\linewidth]{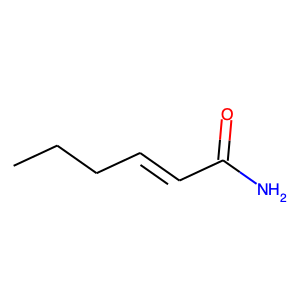}\end{tabular}& \begin{tabular}{l} (2E)-2-Hexenamide\end{tabular} \\ \hline 
\begin{tabular}{l}NCCC1=CC(=O)C1\end{tabular} &  \begin{tabular}{l}\includegraphics[width=0.1\linewidth]{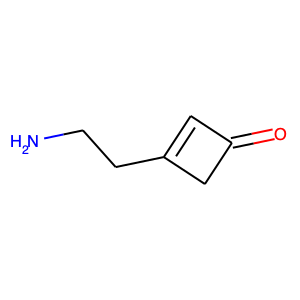}\end{tabular}& \begin{tabular}{l}Multiple search results\end{tabular} \\ \hline 
\begin{tabular}{l}CCC(C)CC(N)=O\end{tabular} &  \begin{tabular}{l}\includegraphics[width=0.1\linewidth]{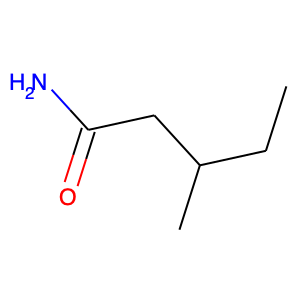}\end{tabular}& \begin{tabular}{l} 3-Methylpentanamide\end{tabular} \\ \hline 
\begin{tabular}{l}NC(=O)C1CCCC1\end{tabular} &  \begin{tabular}{l}\includegraphics[width=0.1\linewidth]{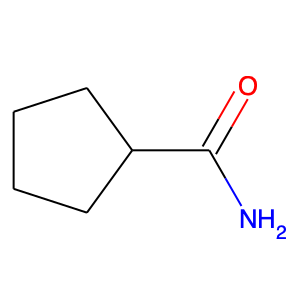}\end{tabular}& \begin{tabular}{l} Cyclopentanecarboxamide\end{tabular} \\ \hline 
\begin{tabular}{l}CCCCC(C)CC\end{tabular} &  \begin{tabular}{l}\includegraphics[width=0.1\linewidth]{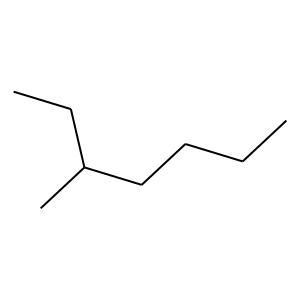}\end{tabular}& \begin{tabular}{l} 3-Methylheptane\end{tabular} \\ \hline 
\begin{tabular}{l}CCCC(O)C(=O)O\end{tabular} &  \begin{tabular}{l}\includegraphics[width=0.1\linewidth]{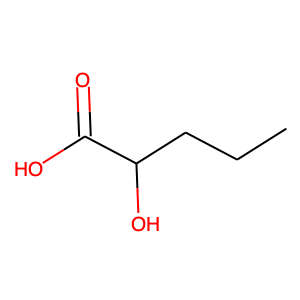}\end{tabular}& \begin{tabular}{l} 2-hydroxyvaleric acid\end{tabular} \\ \hline 
\begin{tabular}{l}O=C1C=CC(O)=CC1\end{tabular} &  \begin{tabular}{l}\includegraphics[width=0.1\linewidth]{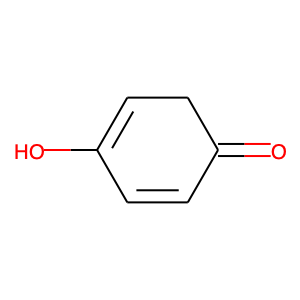}\end{tabular}& \begin{tabular}{l} 4-(2H)Hydroxy-2,4-cyclohexadien-1-one \end{tabular} \\ \hline 
\begin{tabular}{l}O=C(O)C1CC=CC1\end{tabular} &  \begin{tabular}{l}\includegraphics[width=0.1\linewidth]{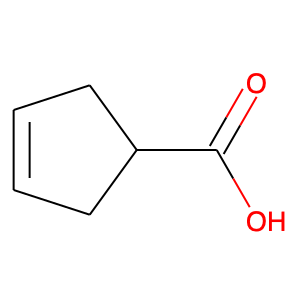}\end{tabular}& \begin{tabular}{l} 3-Cyclopentenecarboxylic Acid\end{tabular} \\ \hline 
\begin{tabular}{l}OCC1CCCCC1\end{tabular} &  \begin{tabular}{l}\includegraphics[width=0.1\linewidth]{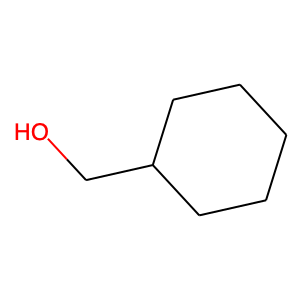}\end{tabular}& \begin{tabular}{l} Cyclohexylmethanol\end{tabular} \\ \hline 
\begin{tabular}{l}CC(CO)CCCN\end{tabular} &  \begin{tabular}{l}\includegraphics[width=0.1\linewidth]{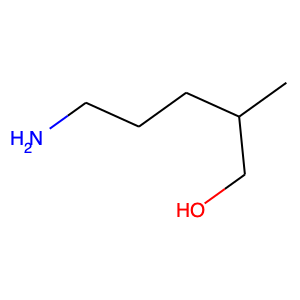}\end{tabular}& \begin{tabular}{l} 5-Amino-2-methyl-1-pentanol\end{tabular} \\ \hline 
\begin{tabular}{l}CC1CCCC(N)N1\end{tabular} &  \begin{tabular}{l}\includegraphics[width=0.1\linewidth]{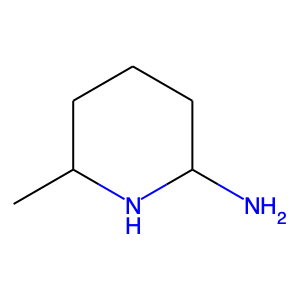}\end{tabular}& \begin{tabular}{l} 6-Methyl-2-piperidinamine\end{tabular} \\ \hline 
\begin{tabular}{l}CCC(=O)CC(N)=O\end{tabular} &  \begin{tabular}{l}\includegraphics[width=0.1\linewidth]{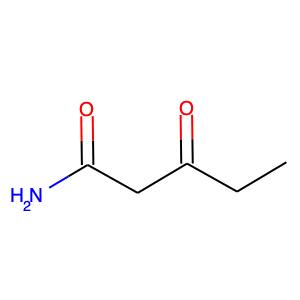}\end{tabular}& \begin{tabular}{l} 3-Oxopentanamide\end{tabular} \\ \hline 
\begin{tabular}{l}CC(O)CCCCN\end{tabular} &  \begin{tabular}{l}\includegraphics[width=0.1\linewidth]{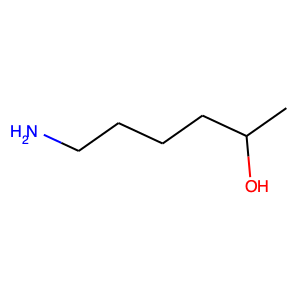}\end{tabular}& \begin{tabular}{l} 6-Amino-2-hexanol\end{tabular} \\ \hline 
\begin{tabular}{l}NC1CC=CC(=O)C1\end{tabular} &  \begin{tabular}{l}\includegraphics[width=0.1\linewidth]{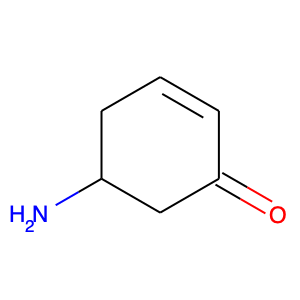}\end{tabular}& \begin{tabular}{l}Multiple search results\end{tabular} \\ \hline 
\begin{tabular}{l}CCNCCCCO\end{tabular} &  \begin{tabular}{l}\includegraphics[width=0.1\linewidth]{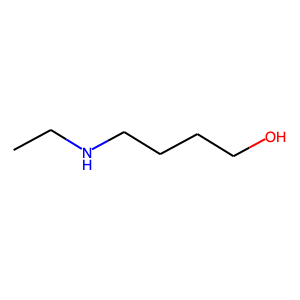}\end{tabular}& \begin{tabular}{l} 4-(Ethylamino)-1-butanol\end{tabular} \\ \hline 
\begin{tabular}{l}NCCCCC(N)=O\end{tabular} &  \begin{tabular}{l}\includegraphics[width=0.1\linewidth]{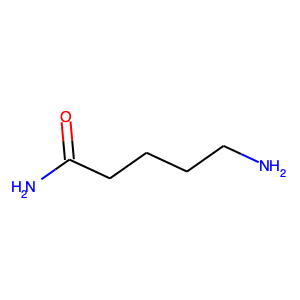}\end{tabular}& \begin{tabular}{l} 5-Aminopentanamide\end{tabular} \\ \hline 
\begin{tabular}{l}CCCCCC(=O)Cl\end{tabular} &  \begin{tabular}{l}\includegraphics[width=0.1\linewidth]{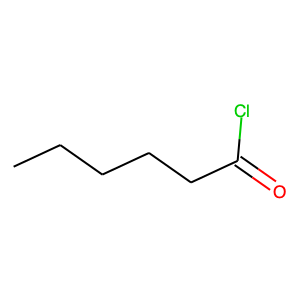}\end{tabular}& \begin{tabular}{l} 506332\end{tabular} \\ \hline 
\begin{tabular}{l}CC(C)CNC(N)=O\end{tabular} &  \begin{tabular}{l}\includegraphics[width=0.1\linewidth]{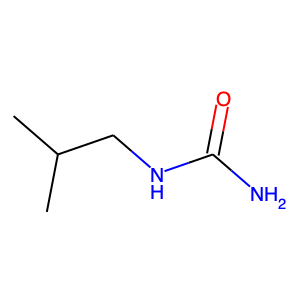}\end{tabular}& \begin{tabular}{l} Isobutylurea\end{tabular} \\ \hline 
\begin{tabular}{l}C=C(C)CCC(=O)O\end{tabular} &  \begin{tabular}{l}\includegraphics[width=0.1\linewidth]{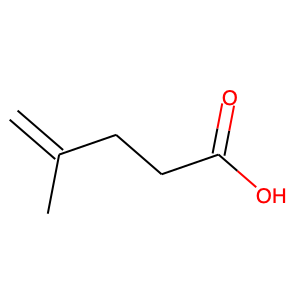}\end{tabular}& \begin{tabular}{l} 4-Methyl-4-pentenoic acid\end{tabular} \\ \hline 
\begin{tabular}{l}OC1=CCCCC1O\end{tabular} &  \begin{tabular}{l}\includegraphics[width=0.1\linewidth]{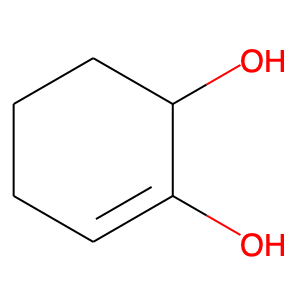}\end{tabular}& \begin{tabular}{l}Multiple search results\end{tabular} \\ \hline 
\end{longtable}

\fi 

\end{document}